\documentclass[11pt,a4paper]{article}
\usepackage[hyperref]{emnlp2020}
\usepackage{times}
\usepackage{latexsym}
\usepackage{graphicx}
\usepackage{url}
\usepackage{amsmath}
\DeclareMathOperator*{\argmax}{arg\,max}

\usepackage{float}
\usepackage{xcolor}
\usepackage{subcaption}
\usepackage{multirow}
\usepackage{amsfonts}
\usepackage{booktabs}
\usepackage{bbm}
\usepackage{pifont}
\usepackage{makecell,rotating}
    \setcellgapes{5pt}

\hypersetup{colorlinks,urlcolor={blue}}
\usepackage{comment}

\usepackage{microtype}

\aclfinalcopy 
\def\aclpaperid{2} 

\title{BERTering RAMS: What and How Much does BERT Already Know About Event Arguments? --- A Study on the RAMS Dataset}
\newenvironment{tight_enumerate}{
\begin{enumerate}
  \setlength{\itemsep}{0pt}
  \setlength{\parskip}{0pt}
}{\end{enumerate}}

\author{Varun Gangal, Eduard Hovy \\
        Language Technologies Institute \\ Carnegie Mellon University \\ {\tt \{vgangal,hovy\}@cs.cmu.edu } }
\date{}

\begin{document}
\maketitle
\begin{abstract}
Using the attention map based probing framework from \cite{clark2019does},  we observe that, on the RAMS dataset \cite{ebner-etal-2020-multi}\footnote{Refer to Figure 1 of that paper for an example illustrating four role names. Since these role names are human readable and intuitively named, we refer to them without elaboration.}, BERT's attention heads\footnote{We use \emph{map} to refer to the per-example word-word activations at a particular layer-head, while \emph{head} refers either to the identity of the particular layer-head. We ground these terms more clearly in \S\ref{subsec:background}} have modest but well above-chance ability to spot event arguments sans \emph{any training or domain finetuning}, varying from a low of 17.77\% for \textit{Place} to a high of  
51.61\% for \textit{Artifact}. Next, we find that linear combinations of these heads, estimated with $\approx$11\% of  available total event argument detection supervision, can push performance well-higher for some roles ---  highest two being \textit{Victim} (68.29\% Accuracy) and \textit{Artifact} (58.82\% Accuracy). Furthermore, we investigate how well our methods do for cross-sentence event arguments. We propose a procedure to isolate ``best heads" for cross-sentence argument detection separately of those for intra-sentence arguments. The heads thus estimated have superior cross-sentence performance compared to their jointly estimated equivalents, albeit only under the unrealistic assumption that we already know the argument is present in another sentence. Lastly, we seek to isolate to what extent our numbers stem from lexical frequency based associations between gold arguments and roles. We propose \textsc{Nonce}, a scheme to create adversarial test examples by replacing gold arguments with randomly generated ``nonce" words. We find that learnt linear combinations are robust to \textsc{Nonce}, though individual best heads can be more sensitive.


\end{abstract}

\section{Introduction}
\label{sec:intro}

The NLP representation paradigm has undergone a drastic change in this decade --- moving from linguistic/task motivated 0-1 feature families to per-word-type pretrained vectors \cite{pennington2014glove} to contextual embeddings \cite{peters2018deep}.

Contextual embeddings (CEs) produce in-context representations for each token - the representation framework being a large, pretrained encoder with per-token outputs. The typical procedure to use CEs for a downstream task is to add one or more task layers atop each token, or for a designated token per-sentence, depending on the nature of the task.

The task layers (and optionally, the representation) are then ``\textit{finetuned}" using a task specific loss, albeit with a slower training rate than would be used for from-scratch training. ELMo \cite{peters2018deep} was an early CE. The three-fold recipe of a transformer based architecture, masked language modelling objective and large  pre-training corpora, starting with BERT \cite{devlin2018bert} led to CEs which were vastly effective for most tasks.

The strong performance of contextual representations with just shallow task layers and minimal finetuning drove the urge to understand what and how much these models \emph{already knew} about aspects of syntax and semantics. The study of methods and analysis to do this has come to be called \emph{probing}. Besides ``explaining" CE featurization, probing can aid in finding lacunae to be addressed by future representations.


\citet{linzen2016assessing}, one of the early works on probing, evaluated whether language models could predict the correct verb form agreeing with the noun. \citet{marvin2018targeted} generalized this approach beyond single-word gaps with a larger suite of ``minimal pairs". They also control for lexical confounding and expand the probing to new aspects such as reflexive anaphora and NPIs. \citet{gulordava2018colorless} evaluate subject-verb agreement but only through ``nonce" sentences to control for both lexical confounding and memorization\footnote{A motivation for our ablation in \S \ref{subsubsec:Confound}}.
\citet{lakretz2019emergence} isolate units of LSTM language models whose activations closely track verb-noun number agreement, particularly for hard, long-distance cases. 
\citet{clark2019does}, whose probing methods we adopt, examine if BERT attention heads capture dependency structure.

\begin{figure*}[!ht]
    \centering
    \includegraphics[width=1.9\columnwidth]{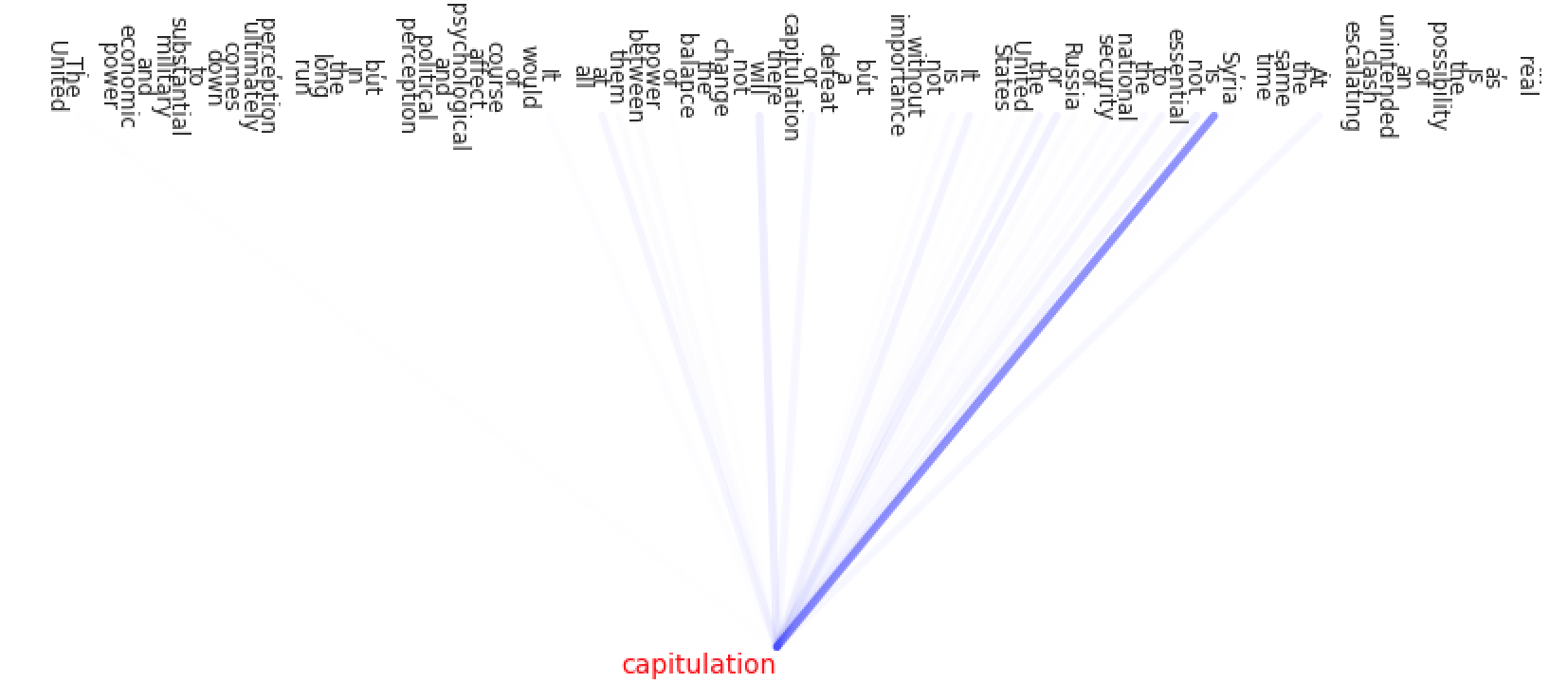}
    \caption{\small In this example, the head chosen by \textsc{BestHead} for the \textsc{Place} role, correctly picks out the argument ``\textit{Syria}" for the trigger ``\textit{capitulation}".
    Attention probabilities are shown as blue lines from trigger token to other tokens, with boldness indicating magnitude. It manages to evade distractor pronouns (\textit{there}) and other geographical entity names (\textit{Russia} and \textit{United States}). The text above flows in right to left direction. The  full text reads: ``\textit{Chances of intentional conflict are real as is the possibility of an unintended clash escalating . At the same time, Syria is not essential to the national security of Russia or the United States. It is not without importance but a defeat or \textbf{capitulation} there will not change the balance of power between them at all \ldots}" }
    \label{fig:place39}
\end{figure*}

In this work, we probe what and how much a pretrained BERT representation already knows about event roles and their arguments. Understanding how well event arguments are represented can be a first foray into understanding other aspects about events. Extraction of event arguments is often a prerequisite for more complex event tasks. Some examples are event coreference \cite{lu2018event}, detecting event-event temporal \cite{vashishtha2019fine}  and causal relations \cite{dunietz2017because}, sub-event structure \cite{araki2014detecting} and generating approximate causal paths \cite{kang2017detecting}. Tuples of event-type and arguments are one way of inducing script like-structures \cite{chambers2008unsupervised}. In summary, our work makes the following contributions:
\begin{tight_enumerate}
    \item We show that there always exists a BERT attention head (\textsc{BestHead}) with above-chance ability to detect arguments, for a given event role. We also show that this ability is even stronger through learnt linear combinations (\textsc{Linear}) of heads.
    \item We notice a relative weakness at detecting cross sentence arguments (\S \ref{subsec:crossSentPerformance}). Motivated by this, we devise a procedure to isolate only the cross-sentence argument detection ability of heads w.r.t a role (\S \ref{subsec:CSO}). Our procedure considerably improves cross-sentence performance for some roles (\S\ref{subsec:crossSentOnlyExperiments}), especially for \textsc{Instrument} and \textsc{Place}.
    \item Lastly, we seek to isolate how much of the zero-shot argument detection ability originates solely from the model's world knowledge and lexical frequency based associations. To do this, we propose \textsc{Nonce}, a method to perturb test examples to dampen such associations (\S \ref{subsubsec:Confound}). We find that the \textsc{Linear} approach is robust to \textsc{Nonce} perturbation, while \textsc{BestHead} is more sensitive. 
\end{tight_enumerate}


\section{Methodology}
\label{sec:methodology}
\subsection{Background}
\label{subsec:background}
\subsubsection{Transformers}
The Transformer architecture \cite{vaswani2017attention} consists of $|L|$ layers, each comprised of $|H|>1$ ``self-attention" heads. Here, we describe the architecture just enough to ground terminology - we defer to the original work for detailed exposition.

In a given layer $l$\footnote{We omit layer index $l$ in the rest of the  passage to declutter  our notation.}, a single self-attention head $h$ consists of three steps - First, query, key and value projections $q^{h}_{i}={Q_{h}}^{T}e_{i},k^{h}_{i}={K_h}^{T}e_{i},v^{h}_{i}={V_h}^{T}e_{i}$ are computed from the previous layer's token embedding $e_{i}$. Then, softmax normalized dot products $\alpha^{h}_{ij} = \frac{(q^{h}_{i})^{T} {k_{j}}^{h}}{\sum_{m} (q^{h}_{i})^{T} {k_{m}}^{h}}$ are computed between the current token's query projection and other token's key projections. These dot products a.k.a \emph{attention values} are then used as weights to combine all token value projections - $o^{h}_i = \sum_{j} \alpha^{h}_{ij} v^{h}_{j} $ gives the current head's token output $o^{h}_i$. Finally, the outputs from all heads are concatenated and projected to get the per-token embeddings for the current layer $o_{i} = W^{T} Concat(\{o^{0}_{i}, o^{1}_{i}\ldots o^{|H|-1}_{i}\})$ 

Henceforth, we refer to the parameter tuple $\{Q_{h,l},K_{h,l},V_{h,l}\}$ , uniquely identified by $h \in \{0,1, \ldots |H|-1\}, l \in \{0,1, \ldots |L|-1\}$ as the ``\textit{attention head}" or simply 
``\textit{head}", while  values $\alpha^{h,l}_{ij}$ are collectively referred to as the ``\textit{attention map}". 

\subsubsection{BERT}
\label{sec:models}
BERT uses a Transformer architecture with 12 heads and 12 layers\footnote{For \emph{bert-base}. \emph{bert-large} uses 24 heads and 24 layers.}. It comes with an associated BPE tokenizer \cite{sennrich2015neural} which tokenizes raw inputs to subwords. Its vocabulary contains three special tokens - [CLS], [SEP] and [MASK]. While [CLS] and [SEP] serve as start and end (or sequence-separator) tokens, [MASK] is used in pretraining as described next.

BERT follows a two-stage pretraining process. In the first stage, also known as masked language modelling (MLM), randomly selected token positions are replaced with [MASK]. The task is to predict the true identities of words at these positions, given the sequence. This stage uses single sentences as training examples. In the second stage, also known as next sentence prediction (NSP), the model is given a pair of sentences (separated by [SEP]), with the task being to predict whether these were truly consecutive or not.  

Unless otherwise mentioned, we use the \textit{bert-base-uncased} model. We use the implementation of BERT from HuggingFace. \footnote{\url{github.com/huggingface/transformers/}} \cite{Wolf2019HuggingFacesTS}

\subsection{Dataset}
\label{sec:datasets}
We use the recently released \textsc{RAMS} dataset \cite{ebner-etal-2020-multi} for all our experiments. The reasons for using this particular dataset for our analysis are
\begin{itemize}
    \item  It has a wide mix of reasonably frequent roles (represented well across splits) from different kinds of frames . Discussion on non-frequent roles can be found in \S \ref{subsec:nonFrequentRoles}.
    \item For many roles, it has examples with the gold arguments being in a different sentence from the event trigger. This makes it easy to probe for intra-sentence and cross-sentence argument extraction in the same set of experiments. Analysis of cross-sentence performance can be found in \S\ref{subsec:crossSentPerformance} and \S\ref{subsec:crossSentOnlyExperiments}
\end{itemize}
We note that the dataset is in English \cite{bender2018data} and observations made may not generalize to other languages.
\subsubsection{Setup}
\label{subsubsec:setup}
For example $x$, we refer to the event, role, gold argument and document as $e$, $r$, $a$ and $D$. $D$ is an ordered sequence of tokens $\{w_{0},w_{1} \ldots w_{|D|-1}\}$. $i_{e}$ denotes the event trigger index\footnote{To simplify our analysis, we do not include multi-word triggers. These form only $\approx$1.6\% of the cases in the dataset.}. 

We use the layer index $l$ and head indices $0$ to $|H|-1$ to index the respective head's attention distribution \emph{from} token $i$ to all other tokens $j \in D$ at index $i_{e}$.  
\vspace{-\abovedisplayskip}
\begin{align}
\label{prob_first}
    P^{*}_{l,h}(j|i_{e}) &= \alpha^{i_{e}j}_{l,h}, 0\leq l < |L|, 0 \leq h < |H|
\end{align} 
Note, however that there exist a complementary set of attention values from each token $j$ \emph{to} the token $i_{e}$. To use a unified indexing scheme to refer to these values, we use negative indices from $-1$ to $-|H|$ as their head indices. Since these values come from attention-head activations of different positions, they need to be renormalized to use them as probabilities.
\vspace{-\abovedisplayskip}
\begin{align}
\label{prob_second}
   P^{*}_{l,-h}(j|i_{e}) &= \frac{\alpha^{j{i_{e}}}_{l,h-1}}{\sum_{k \in D} \alpha^{k{i_e}}_{l,h-1}}, 0 < h \leq |H|
\end{align} 

\subsubsection{Words and Subwords}
\label{subsubsec:wordsandsubwords}
Our above framework assumed that the attention maps are between whole word tokens. However, \textsc{Bert} represents a sentence as a sequence of BPE-subwords at every level, including for the attention maps. 

We use the quite intuitive approach described in Section 4.1 of \cite{clark2019does} - incoming attentions to constitutent subwords of a word are added to get the attention to that word. Outgoing attention values from constituent subwords are averaged to get the outgoing attention value from the word.

Note that the above operations precede the probability computations in Equations \ref{prob_first} and \ref{prob_second}.

\subsubsection{Dataset Splits}
We follow the practice of earlier probing works such as \cite{sorodoc2020probing} and \cite{linzen2016assessing} of using one of the smaller splits for training. Specifically, we use the original dev split of RAMS ($924$ examples in total) as our training split. Note that each example could contain multiple role-argument pairs.

\begin{table}[!ht]
    \centering
    \small
    \begin{tabular}{c|c|c}
        \toprule
        Splits &  Examples & Tokens\\
        \midrule
        Train (Original Dev) & 924 & 0.12M  \\
        Dev (Original Test) & 871 & 0.11M  \\
        Test (Original Train) & 7329 & 0.98M \\
        \bottomrule
    \end{tabular}
    \caption{\small Split example counts and token sizes from the \textsc{RAMS}. Note that we use different splits since our work is a probing exercise.}
    \label{tab:diffculty_corr}
\end{table}

\subsection{Evaluation Measure}
\label{subsec:evalmeasures}
For a given event $e$ and role $r$, we define a predicted argument token index $\widehat{a}$ to be accurate if it corresponds to any of the tokens in the gold argument span $[a_{r,e}^{beg},a_{r,e}^{end}]$. This is  described formally in Equation \ref{eq:accuracy}. ${I}$ stands for the 0-1 indicator function.
\vspace{-\abovedisplayskip}
\begin{align}
    Acc_{e,r,a}(\widehat{a}) &=  {I}( a_{r,e}^{beg} \leq \widehat{a} < a_{r,e}^{end})
    \label{eq:accuracy}
\end{align}
Typical measures of argument extraction differ from the one we use, being span-based. Given the limitations of our probing approaches, we lack a clear mechanism of predicting multi-word spans, and can only predict likely single tokens for the argument, which led us to choose this measure\footnote{We will interchangeably refer to $Acc$ as just ``\textit{accuracy}" in plain-text in the rest of the paper}.

\subsection{Approaches}
\label{subsec:approaches}
\subsubsection{\textsc{BestHead}}
Let $X = \{e_m,r_m,a_m\}_{m=1}^{m=M}$ be the training set. $X_{r}$ is the subset of training examples with $r_{m}=r$. For each role $r$, \textsc{BestHead} selects the head ${\{l,h\}}_{best}(r)$ with best aggregate accuracy on $X_{r}$. Other than one pass over the training set for comparing aggregate accuracies of heads for each role, there is no learning required for this method. At test-time, based on the test role, the respective best head is used to predict the argument token.
\vspace{-0.7\abovedisplayskip}
\par\nobreak{\small
\begin{align*}
  {Acc}^{X_{r}}_{l,h}   &= \sum_{m=1}^{m=M_{r}} Acc_{e_m,r,a_m}(\argmax_{j} \widehat{P}_{l,h}(j|i_{e_m})) \\
 {\{l,h\}}_{best}(r)    &= \argmax_{l,h} Acc^{X_{r}}_{l,h}
\end{align*}
\normalsize}%
\subsubsection{\textsc{Linear}}
The \textsc{Linear} model learns a  weighted linear combination of all $|L| \times |H| \times 2$ head distributions (twice for the ``\textit{from}" and "\textit{to}" heads). 
\begin{align*}
    \phi(j|i) &= \sum_{l=0}^{l=|L|-1} \sum_{h=-|H|}^{h=|H|-1} w_{l,h} \widehat{P}_{l,h}(j|i) + B \\
    \widehat{P}(j|i) &= \frac{\phi(j|i)}{\sum_{k=0}^{k=|D|-1} \phi(k|i)}
\end{align*}
Note that gradients are not backpropagated into BERT - only the linear layer parameters $w_{l,h},B$ are updated during backpropagation. This formulation is the same as the  one in \cite{clark2019does}.

For our loss function, we use the KL Divergence $KL(\widehat{P}||P)$ between the predicted distribution over document tokens $\widehat{P}$ and the gold distribution over document tokens $P$. For the gold distribution over arg tokens, the probability mass is equally distributed tokens in the argument span, with zero mass on the other tokens.
\vspace{-\abovedisplayskip}
\begin{align*}
KL(\widehat{P}||P)&= \sum_{k=0}^{k=|D|-1} \widehat{P}(k|i) \log{\frac{\widehat{P}(k|i)}{P(k|i)}}
\end{align*}

\subsection{Baselines}

\subsubsection{\textsc{Rand}}
The expected accuracy of following the strategy of randomly picking any token $i$ from the document $D$ as the argument (other than the trigger word $i_e$ itself). For a given role $r$ with a gold argument $a_{r,e}$ of length $|a_{r,e}|$, this equals $\frac{|a_{r,e}|}{|D|-1}$.

\subsubsection{\textsc{SentOnly}}
The expected accuracy of following the strategy of randomly picking any token from the same sentence $S_e$ as the argument, save the event trigger itself. This is motivated by the intuition that event arguments mostly lie in-sentence. This equals $\frac{|a_{r,e}|}{|S_{e}|-1}$


\subsubsection{\textsc{Nonce} procedure}
\label{subsubsec:Confound}
We wish to isolate how much of the heads performance is due to memorized ``world knowledge" and typical lexical associations e.g \textit{Russia} would typically always be a \textsc{Place} or \textsc{Target}. Recent works have shown that BERT does retain such associations, including for first names \cite{shwartz2020you}, and enough so that it can act as a reasonable knowledge base \cite{petroni2019language}.

One way of implementing this is  to create  perturbed test examples where gold arguments are replaced with synthetically created ``nonce" words not necessarily related to the context. This is similar to the approach of \cite{gulordava2018colorless}.

\begin{itemize}
    \item Each gold argument token is replaced by a randomly generated token with the  same number of characters as the original string.
    \item Stop words such as determiners, pronouns, and conjunctions are left unaltered, though they might be a part of the argument span.
    \item We also ensure that the shape of the original argument, i.e the profile of case, digit vs letter is maintained\footnote{We are aware that case mostly doesn't matter since we use bert-*-uncased in most experiments}. e.g \textit{Russia-15} can be randomly replaced by \textit{Vanjia-24}, which has the same shape \textit{Xxxx-dd}.
    \item Note that we do not take pronounceability of the nonce word into account. Though this could arguably be a relevant invariant to maintain, we were not sure of an apt way to enforce it automatically.
    \item We also note that BERT may end up using a likely larger number of subword tokens to replace the nonce words than it would use for the gold argument token. Since these are essentially randomly composed tokens, they can contain subwords which are rarely seen in vocabulary tokens.
\end{itemize}

We refer to this procedure as \textsc{Nonce}, and overloading the term, the test set so created as the \textsc{Nonce} test set.

\section{Experiments}
\label{sec:experiments}
\subsection{Spotting the Best Head}
\label{subsec:bestHead}
\begin{table}
    \centering
    \scriptsize
    \begin{tabular}{l|c|c}
    \toprule
       Role  & ${\text{\{l,h\}}}_{best}$ & \%Accuracy      \\
         \midrule
        \textsc{Defendant} & 8,10 & 35.90   \\
        \textsc{Destination} & 0,8 & 21.43   \\
        \textsc{Origin} & 7,-1 & 31.82  \\
        \textsc{Transporter} & 8,10 & 31.58  \\
        \textsc{Instrument} & 9,7 & 31.37  \\
        \textsc{Beneficiary} & 8,10 & 26.56   \\
        \textsc{Attacker} & 7,-8 & 33.93  \\
        \textsc{Target} & 9,1 & \textbf{44.61}   \\
        \textsc{Giver} & 8,10 & 25.55  \\
        \textsc{Victim} & 9,1 & \textbf{46.34}  \\
        \textsc{Artifact} & 4,-6 & \textbf{50.42}  \\
        \textsc{Communicator} & 8,10 & \textbf{51.61}  \\
       \textsc{Participant} & 8,10 & 28.57   \\
        \textsc{Recipient} & 7,10 & \textbf{40.78}  \\
        \textsc{Place} & 9,1 & 17.77  \\        
        \bottomrule
    \end{tabular}
    \caption{\small Best layer-head pair , ${\text{\{l,h\}}}_{best}$ and \% Accuracy  for the 15 most frequent roles in RAMS, using \textit{bert-base-uncased}. +ve h indices denote ``from" heads, while -ve indices denote ``to" heads, as explained in \S\ref{subsubsec:setup}}
    \label{tab:bestHeadRes}
\end{table}
In Table \ref{tab:bestHeadRes}, we record the accuracies and layer positions of best heads for the 15 most frequent roles.
\begin{tight_enumerate}
\item \textsc{BestHead} always has higher accuracy than the \textsc{Rand} and \textsc{Sent} baselines.
\item 5 of the 15 roles can be identified with 40\%+ accuracy - the highest being  \textsc{Communicator} , at 51.61\%.
\item The best head for arguments which are not together present in frames is often the same. For instance, Layer 8, Head 10 is the best head for \textsc{Transporter}, \textsc{Attacker}, \textsc{Communicator} and \textsc{Beneficiary}.
\item  Most best heads are located in the higher layers, specifically the 7th, 8th or 9th layers. An exception are the best head for \textsc{Destination} and \textsc{Artifact} roles,  located in the 0th layer and 4th layers respectively.
\item  \textit{Place} roles are the hardest to identify, with an accuracy of $17.77\%$.
\item  Layer 8, Head 10 seems to be doing a lot of the heavylifting. For 7 out of 15 roles, this is the best head. This shows that it is  quite ``overworked" in terms of the number of roles it tracks. Furthermore, though some of these role pairs are from different frames (e.g see Point 3 above), some aren't, e.g \textsc{Giver} and \textsc{Beneficiary}. In such cases, atleast one of the two arguments predicted for these two roles is sure to be inaccurate - e.g the head would point to either the \textsc{Giver} or \textsc{Beneficiary}, but not both. \footnote{It is quite non-intuitive for \textsc{Giver} and \textsc{Beneficiary} spans to overlap --- we don't see any examples with the same.}
\item Most of the best heads for roles are ``from" heads rather than ``to" heads, apart from those for \textsc{Origin}, \textsc{Attacker} and \textsc{Artifact}.
\end{tight_enumerate}

\subsection{\textsc{Linear} Performance}
\label{subsec:LinearExperiments}

\begin{table}
    \centering
    \scriptsize
    \scalebox{0.95}{
    \begin{tabular}{l|c|c|c|c}
    \toprule
       Role  & \textsc{Rand} &  \textsc{SentOnly} &  \textsc{BestHead} &  \textsc{Linear}  \\
         \midrule
        \textsc{Defendant} & 1.75 & 6.98 & 35.90 & 56.41   \\
        \textsc{Destination} & 1.91 & 7.67  & 21.43 & 39.28   \\
        \textsc{Origin} & 1.36  & 7.27   & 31.82 & 28.79 \\
        \textsc{Transporter} & 1.63  & 6.57  & 31.58 & 43.42 \\
        \textsc{Instrument} & 1.88 & 6.29   & 31.37 & 25.49  \\
        \textsc{Beneficiary} & 1.34  & 6.28    & 26.56 & 34.37  \\
        \textsc{Attacker} & 2.07 & 8.52   & 33.93 & 46.43 \\
        \textsc{Target} & 1.78  & 7.30   & 44.61 & 44.61  \\
        \textsc{Giver} & 1.58  & 6.29    & 25.55 & 32.22 \\
        \textsc{Victim} & 1.50  & 6.42   & 46.34 & 68.29 \\
        \textsc{Artifact} & 1.86  & 7.62   & 50.42 & 58.82  \\
        \textsc{Communicator} & 1.58  & 6.55    & 51.61 & 63.71 \\
       \textsc{Participant} & 1.49  & 6.19    & 28.57 & 30.72  \\
        \textsc{Recipient} & 1.83 & 8.57     & 40.78 & 44.69 \\
        \textsc{Place} & 1.67 & 6.84  & 17.77 & 31.93  \\        
        \bottomrule
    \end{tabular}}
    \caption{\small Test accuracies using all the baselines and probe approaches described in \S\S \ref{subsec:approaches} for the 15 most frequent roles in \textsc{RAMS}. Both \textsc{BestHead} and \textsc{Linear} probes outdo the baselines. \textsc{Linear} usually does better, but not for all roles (e.g \textsc{Origin}). Refer to \S\ref{linear_combinations} for a longer discussion.}
    \label{tab:allApproaches}
\end{table}

Table \ref{tab:allApproaches} shows test accuracies for both \textsc{Linear} and \textsc{BestHead} approaches, and also the baselines.

For 12 of the 15 roles, \textsc{Linear} has higher accuracy than \textsc{BestHead}. There are three exceptions - \textsc{Origin} and  \textsc{Instrument}, which suffer a decline and \textsc{Target}, which remains the same. A possible reason could be the higher fraction of cross-sentence gold arguments for these roles. The five roles with lowest number of intra-sentence arguments are \textsc{Destination} (58.92\%), \textsc{Instrument} (62.74\%),  \textsc{Origin} (68.18\%), \textsc{Place} (70.48\%) and \textsc{Target} (81.53\%). 

While \textsc{Destination} and \textsc{Place} do see increases in \textsc{Linear} compared to \textsc{BestHead}, it could be the case that none of the individual heads are particularly good at capturing cross-sentence arguments for the other three roles, while the best head is already good enough to capture the intra-sentence case. This would make \textsc{Linear} not any more rich as a hypothesis space compared to \textsc{BestHead} - causing the similar or slightly worse accuracy. In \S \ref{subsec:crossSentPerformance} we dig deeper into the aspect of cross-sentence performance.
\label{linear_combinations}

\subsection{Cross-Sentence Performance}
\label{subsec:crossSentPerformance}
From Table \ref{tab:crossSentenceOnlyApproaches}, we observe that both \textsc{BestHead} and \textsc{Linear} performance degrades in the cross-sentence case i.e when ``trigger sentence" and ``gold argument sentence" differ. Three potential reasons:
\begin{tight_enumerate}
    \item There are too few instances of cross-sentence event arguments in the small supervised set we use. Furthermore, even if there are a sufficient quantum of cross-sentence event arguments, these form a much smaller proportion of the total instances in comparison to the intra-sentence instances.
    \item Because the limited number of attention heads are already dominated by intra-sentence aspects such as dependency relations, punctuation and subject-verb agreement \cite{clark2019does}, it is difficult for a single attention head to have a higher value for outside sentence tokens compared to in-sentence ones.
    \item Different heads might be best for intra and cross-sentence performance, and finding one best head for both could be sub-optimal.
\end{tight_enumerate}

\subsubsection{Cross Sentence Occlusion (\textsc{CSO})}
\label{subsec:CSO}

Motivated by the above reasons, we devise a  procedure which we refer to as cross-sentence occlusion (\textsc{CSO}). Since Reason 1 is a property of the data distribution, we attempt to alleviate Reasons 2 and 3. To address Reason 3, we try to learn a different head (combination) for the cross-sentence case. To address Reason 2, while finding the best cross-sentence head, we zero-mask out the attention values corresponding to in-sentence\footnote{RAMS comes with a given sentence segmentation.} tokens and re-normalize the probability distribution. 

In practice, one would not be able to use two separate argument detectors for the intra and cross-sentence cases for the same role, since ground-truth information of whether the argument is cross-sentence would be unavailable. We assume this contrived setting only to allow easy analysis\footnote{And also so that we can validate our diagnosis for poor cross-sentence performance in \S 3.3}, and to gloss over the lack of an intuitive zero-shot mechanism of switching between the two cases, when predicting arguments using just attention heads.

\subsection{\textsc{+CSO} Results}
\label{subsec:crossSentOnlyExperiments}


\begin{table}
    \centering
    \scriptsize
    \begin{tabular}{l|l|l}
    \toprule
       Role   &  \textsc{BestHead+CSO} &  \textsc{Linear+CSO}   \\
         \midrule
        \textsc{Origin}      & 4.76 (31.82$\rightarrow$0.00) & 4.76 (56.41$\rightarrow$16.34) \\
        \textsc{Instrument}    & 47.37 (31.37$\rightarrow$21.22) & 52.63 (25.49$\rightarrow$31.51)  \\
       \textsc{Participant}       & 24.99 (28.57$\rightarrow$6.37) & 29.16 (30.72$\rightarrow$6.37)  \\
        \textsc{Place}    & 15.31 (17.77$\rightarrow$10.18) & 30.61 (31.93$\rightarrow$9.30)  \\        
        \bottomrule
    \end{tabular}
    \normalsize
    \caption{\small Accuracies on  cross-sentence test examples using \textsc{BestHead+CSO} and \textsc{Linear+CSO}. The values $Acc_{Total}$$\rightarrow$$Acc_{Cross}$ in parentheses are the total test accuracy and cross-sentence test accuracy respectively, using the simple version of the same approach i.e \textsc{BestHead} and \textsc{Linear}. The \% of cross-sentence examples for each role are: \{\textsc{Origin}:31.82 \textsc{Instrument}:37.26 \textsc{Participant}:17.14 \textsc{Place}:29.52\}}
    \label{tab:crossSentenceOnlyApproaches}
\end{table}
From Table \ref{tab:crossSentenceOnlyApproaches},  we can observe the improvement in cross-sentence test accuracy when using the $+\textsc{CSO}$ approach over its simple counterpart, both for \textsc{BestHead} and \textsc{Linear}. 
The only exception to this is the \textsc{Origin} role, where \textsc{Linear} betters \textsc{Linear-CSO}.

For the \textsc{Instrument} role, both \textsc{BestHead+CSO} and \textsc{Linear+CSO} get close to $50\%$ accuracy. In part, their relatively stronger performance can be explained by \textsc{BestHead} and \textsc{Linear} already being relatively better at detecting cross-sentence \textsc{Instrument} (just above 20\%, but higher than the sub-15 accuracies on the other roles). Nevertheless, \textsc{CSO} still leads to a doubling of accuracies for both approaches. 

We highlight here again that these numbers are only on that subset of the test set where we know that the gold arguments are located in other sentences - though this setting is useful for analysis, a model actually solving this task won't have access to this information. 

Even in our case, there is no obvious way to have a consolidated probe which uses a \textsc{Linear+CSO} and \textsc{Linear} component together, since this would require learning an additional component which predicts whether the gold arguments lie intra-sentence or across-sentence.

\subsection{Effect of \textsc{Nonce}}
\begin{figure}[ht]
\begin{subfigure}{\columnwidth}
    \centering
    \includegraphics[width=0.95\columnwidth]{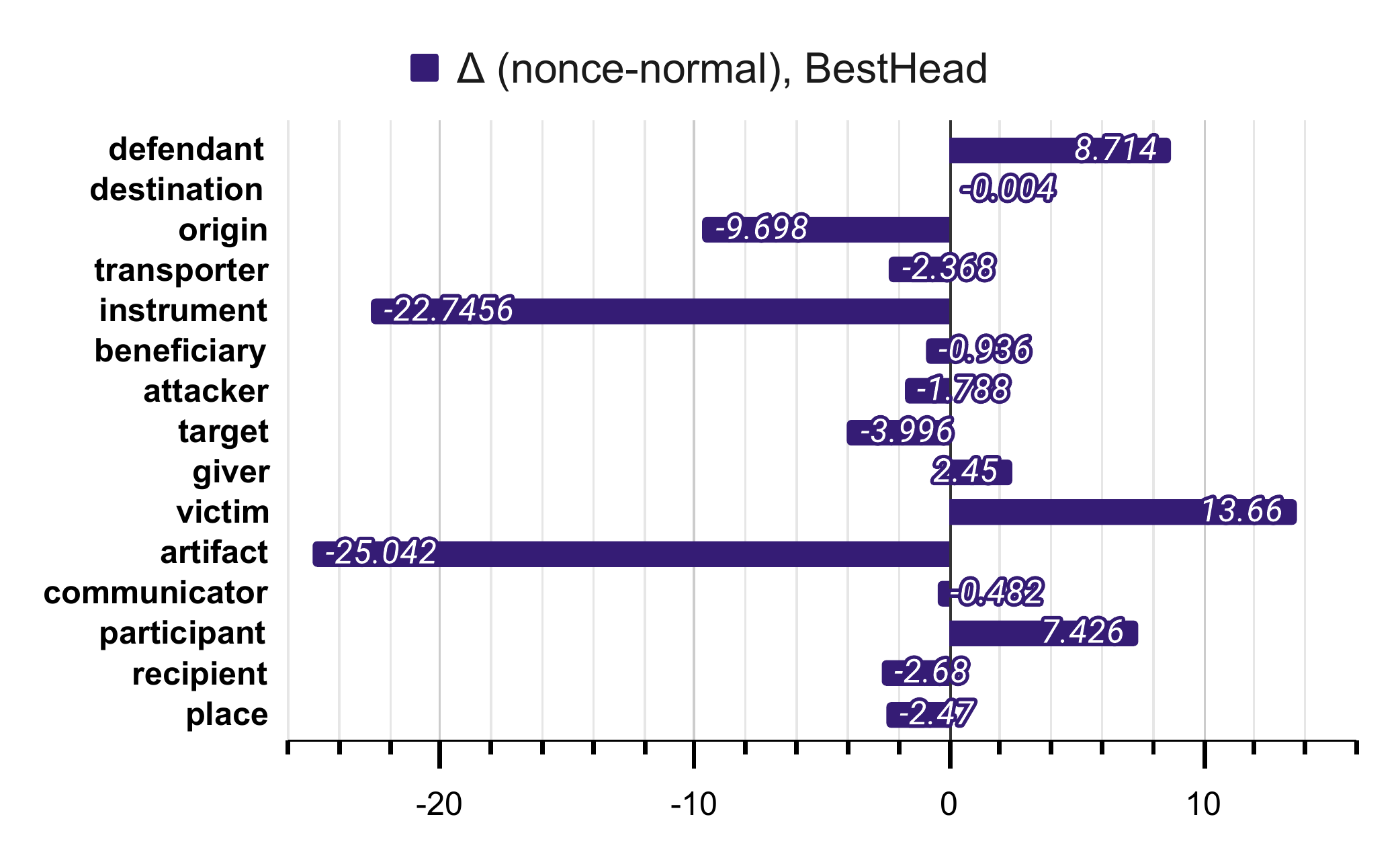}
    \caption{\small \textsc{BestHead}}
    \label{fig:BestHeadConfound}
\end{subfigure} \\
\begin{subfigure}{\columnwidth}
    \centering
    \includegraphics[width=0.95\columnwidth]{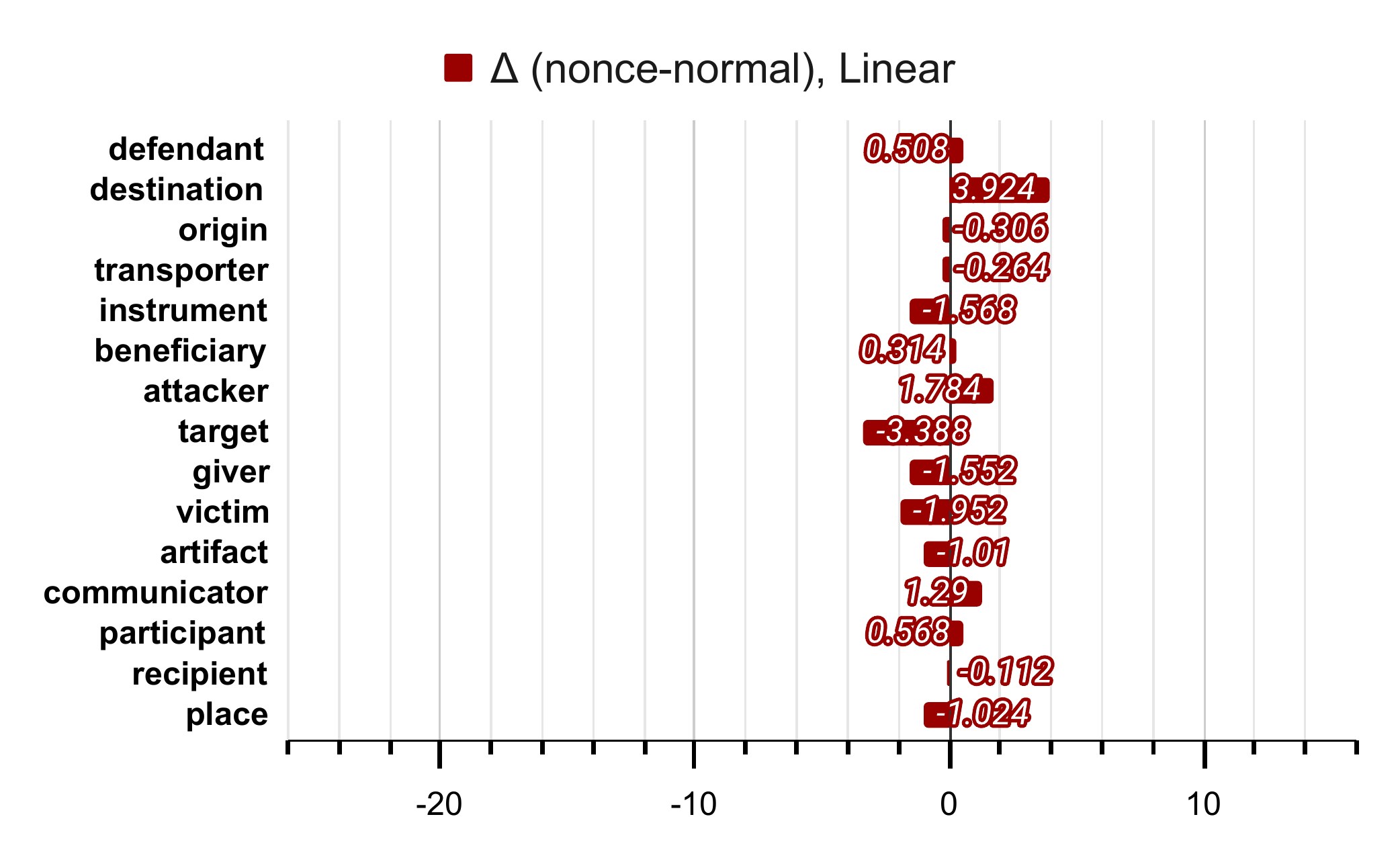}
    \caption{\small \textsc{Linear}}
    \label{fig:LinearConfound}
\end{subfigure}
\caption{Difference in a) \textsc{BestHead} and b) \textsc{Linear} accuracies over normal and \textsc{Nonce} test sets}
\end{figure}

In Figures \ref{fig:BestHeadConfound} and \ref{fig:LinearConfound}, we compare the performances of our methods on perturbations of the test set created using the \textsc{Nonce} procedure outlined in \S \ref{subsubsec:Confound} with their normal test performance. Since \textsc{Nonce} is stochaistic, corresponding results are averaged over \textsc{Nonce} sets created with 5 different seeds.

\textsc{BestHead} test performance is more sensitive to \textsc{Nonce} than \textsc{Linear}. Especially for \textsc{Instrument}, \textsc{Artifact} and \textsc{Origin}, the decrease in accuracy is quite drastic. Surprisingly, we also see  increases for 4 of the 15 roles - \textsc{Defendant}, \textsc{Giver}, \textsc{Victim} and \textsc{Participant}. All other roles see small decreases. For \textsc{Linear}, however, most roles are largely unmoved by \textsc{Nonce}, showing that \textsc{Linear} relies less on lexical associations.

\subsection{Non-Frequent Roles}
\label{subsec:nonFrequentRoles}
So far, we've focussed on analyzing the 15 most frequent roles. In this subsection, we also evaluate our approaches for some non-frequent roles outside this set, such as \textsc{Preventer} and \textsc{Prosecutor}. The results are presented in Table \ref{tab:allApproachesNonFrequentRoles}. Note that, owing to high sparsity for these roles, these results should be taken with ``a pinch of salt" (which is why we chose to separate them out from the frequent roles).

For the frequent roles, we had seen that \textsc{Linear} was mostly better than, or equally good as \textsc{BestHead}. For the non-frequent roles, we see that the comparative performance of \textsc{Linear} vs \textsc{BestHead} varies a lot more - \textsc{Linear} is better for 6 of the 11 roles, and worse for the other 5. The fall in \textsc{Linear} performance is largest for \textsc{Prosecutor} (58.33 $\rightarrow$ 16.67).

We conjecture that this drop is due to poor generalization  as a result of learning from lesser supervision as a result of the roles being non-frequent.
Since \textsc{BestHead} has only two parameters (identity of the best head) compared to the $289$ parameters of \textsc{Linear}, the latter is more sensitive to this problem.

Secondly, we notice that the gap between \textsc{BestHead} and the \textsc{Rand} and \textsc{SentOnly} baselines is much narrower. For \textsc{Vehicle} and \textsc{Money},  \textsc{SentOnly} even outdoes \textsc{BestHead}. For \textsc{Vehicle}, the \textsc{BestHead} accuracy even drops to 0. However, in all these cases, we find that \textsc{Linear} still manages to outdo both baselines. We conjecture that these cases could be due to the best head predicted not being very generalizable due to  small training set size (for that role). Though \textsc{Linear} would also suffer from poor generalization in this case, it might stand its ground better since it relies on multiple heads rather than just one.

\begin{table}
    \centering
    \tiny
    \begin{tabular}{l|c|c|c|c}
    \toprule
       Role  & \textsc{Rand} &  \textsc{SentOnly} &  \textsc{BestHead} &  \textsc{Linear}  \\
         \midrule
        \textsc{Preventer} & 1.51 & 7.30 & 15.00 & 43.14   \\
        \textsc{Passenger} & 1.46 & 6.79  & 57.45 & 45.45   \\
        \textsc{Crime} & 3.06  & 12.62 & 25.81 & 57.99 \\
        \textsc{Injurer} & 1.52  & 6.51  & 29.03 & 16.13\\
        \textsc{Employee} & 1.47 & 6.87   & 53.85 & 50.00  \\
        \textsc{Killer} & 1.75  & 10.63    & 14.29 & 47.62 \\
        \textsc{Money} & 1.51  & 7.77   & 4.17 & 25.00 \\
        \textsc{Detainee} & 1.60  & 8.39  & 62.50 & 50.00  \\
        \textsc{Prosecutor} & 1.55  & 6.34 & 58.33 & 16.67 \\
       \textsc{JudgeCourt} & 1.33  & 6.78  & 22.22 & 41.67  \\
        \textsc{Vehicle} & 1.47 & 6.15  & 0 & 18.18 \\
        \bottomrule
    \end{tabular}
    \caption{\small Test accuracies using all the baselines and probe approaches described in \S\S \ref{subsec:approaches} for some non frequent roles. Both \textsc{BestHead} and \textsc{Linear} probes still outdo the baselines in most cases, but not as convincingly as for frequent roles. Unlike the frequent roles case, \textsc{Linear} actually does worse than \textsc{BestHead} for many roles.}
    \label{tab:allApproachesNonFrequentRoles}
\end{table}

\subsection{Cased vs Uncased}
\label{subsec:casedvsuncased}
\begin{figure}[ht]
\begin{subfigure}{0.99\columnwidth}
    \centering
    \includegraphics[width=0.95\columnwidth]{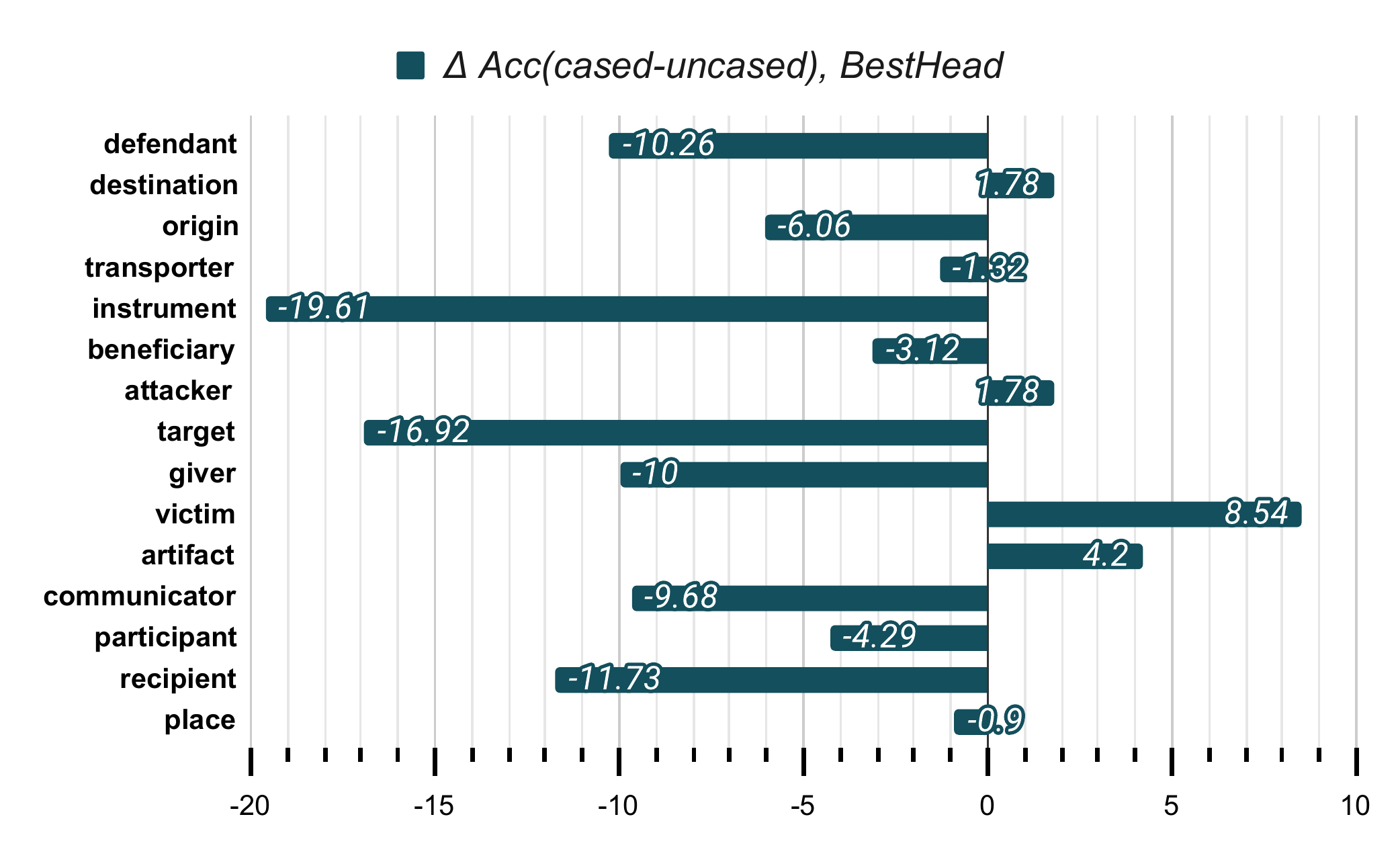}
    \caption{\textsc{BestHead}}
    \label{fig:casedvsuncased}
\end{subfigure} \\
\begin{subfigure}{0.99\columnwidth}
    \centering
    \includegraphics[width=0.95\columnwidth]{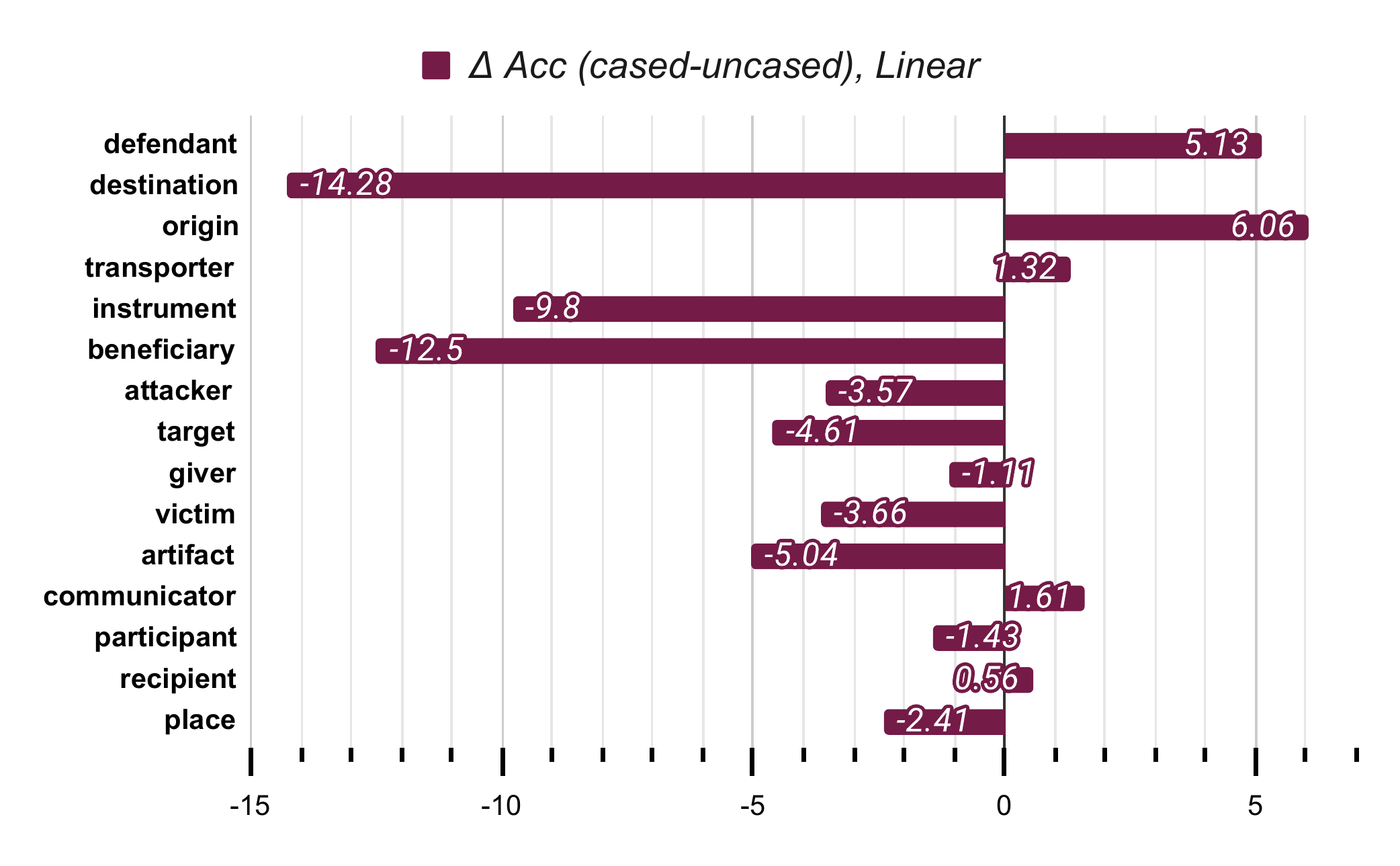}
    \caption{\textsc{Linear}}
    \label{fig:casedvsuncasedlinear}
\end{subfigure}
\caption{$\Delta$ in Test accuracy of a) \textsc{BestHead} b) \textsc{Linear} using \textit{bert-base-uncased} vs \textit{bert-base-cased}}
\end{figure}
In our analysis so far, we have been using the same contextual embedding mechanism throughout, namely \textit{bert-base-uncased}. In Figure \ref{fig:casedvsuncased}, we plot the difference of \textsc{BestHead} test accuracies when using \textit{bert-base-cased} vs \textit{bert-base-uncased}. 

We can see that \textit{bert-base-uncased} is better for most roles - except for \textit{Attacker}, \textit{Victim} and \textit{Artifact}. We also notice that the best layer-head configuration  $\{l_{best},h_{best}\}$ is mostly not preserved between the \textit{bert-base-cased} and \textit{bert-base-uncased} scenarios.
The difference between \textit{bert-base-uncased} and \textit{bert-base-cased} is even more drastic in the cross sentence only experiment , for instance, while there exists a single head which can find cross-sentence \textit{Instrument} args with 37\% accuracy, the best such head of \textit{bert-base-cased} has only 17\% accuracy.


\subsection{Qualitative Examples}
In Figure \ref{fig:qualExamples}, we illustrate some examples of \textsc{BestHead} identifying arguments. 
We defer further discussion to Appendix \S A owing to lack of space.
\begin{figure}[ht]
\begin{subfigure}{.99\columnwidth}
    \centering
    \includegraphics[width=0.99\columnwidth,height=0.49\columnwidth]{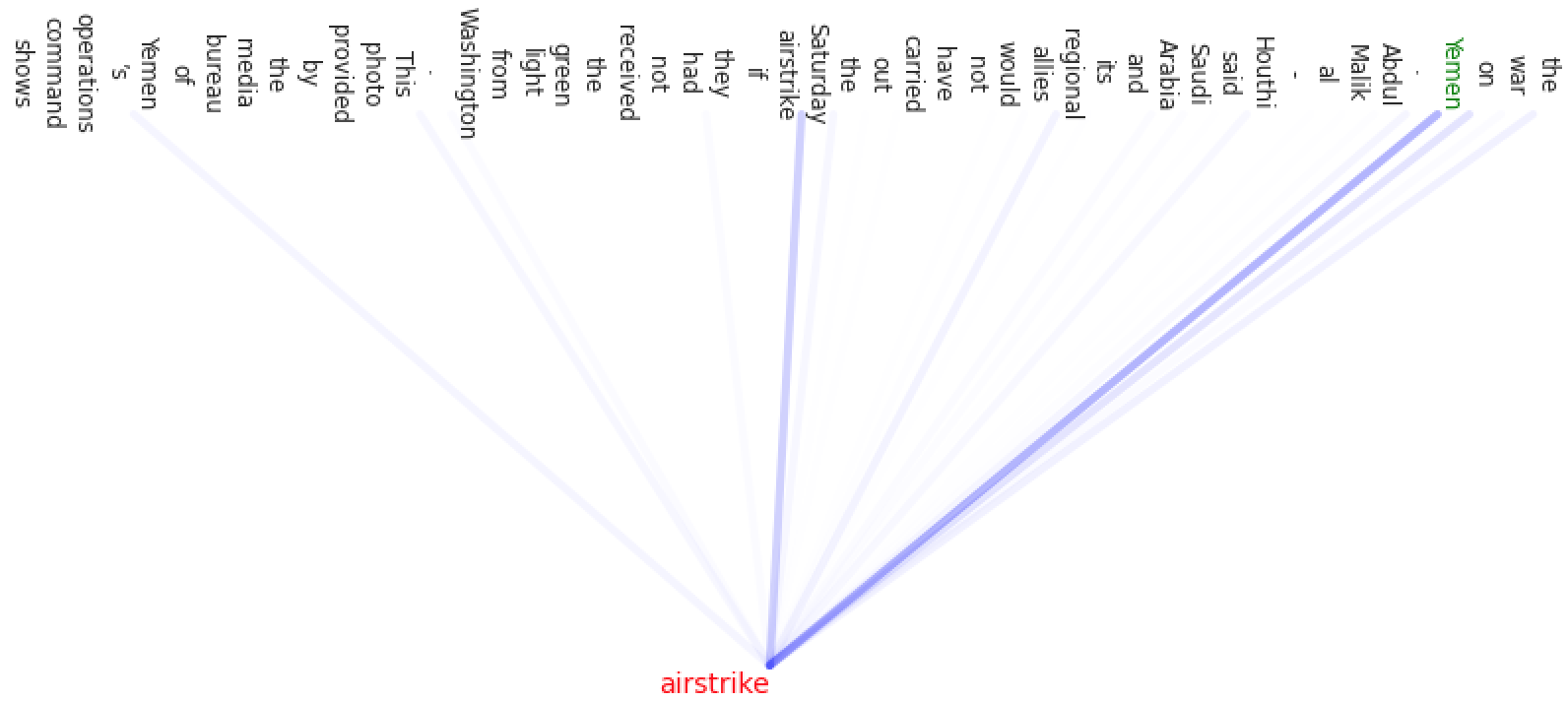}
    \caption{}
    \label{fig:target103}
\end{subfigure} \\
\begin{subfigure}{.99\columnwidth}
    \centering
    \includegraphics[width=0.99\columnwidth,height=0.49\columnwidth]{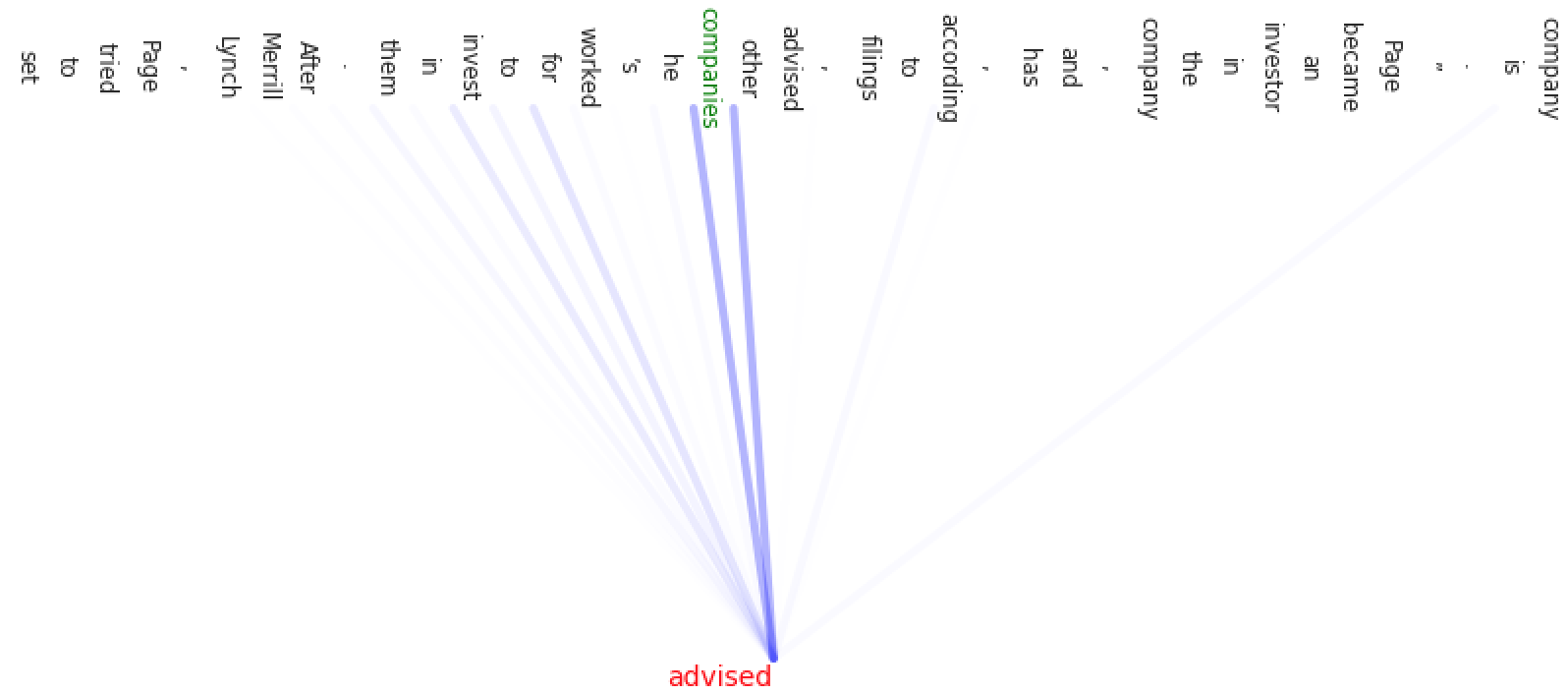}
    \caption{}
    \label{fig:recipient133}
\end{subfigure}
\begin{subfigure}{.99\columnwidth}
    \centering
    \includegraphics[width=0.99\columnwidth,height=0.49\columnwidth]{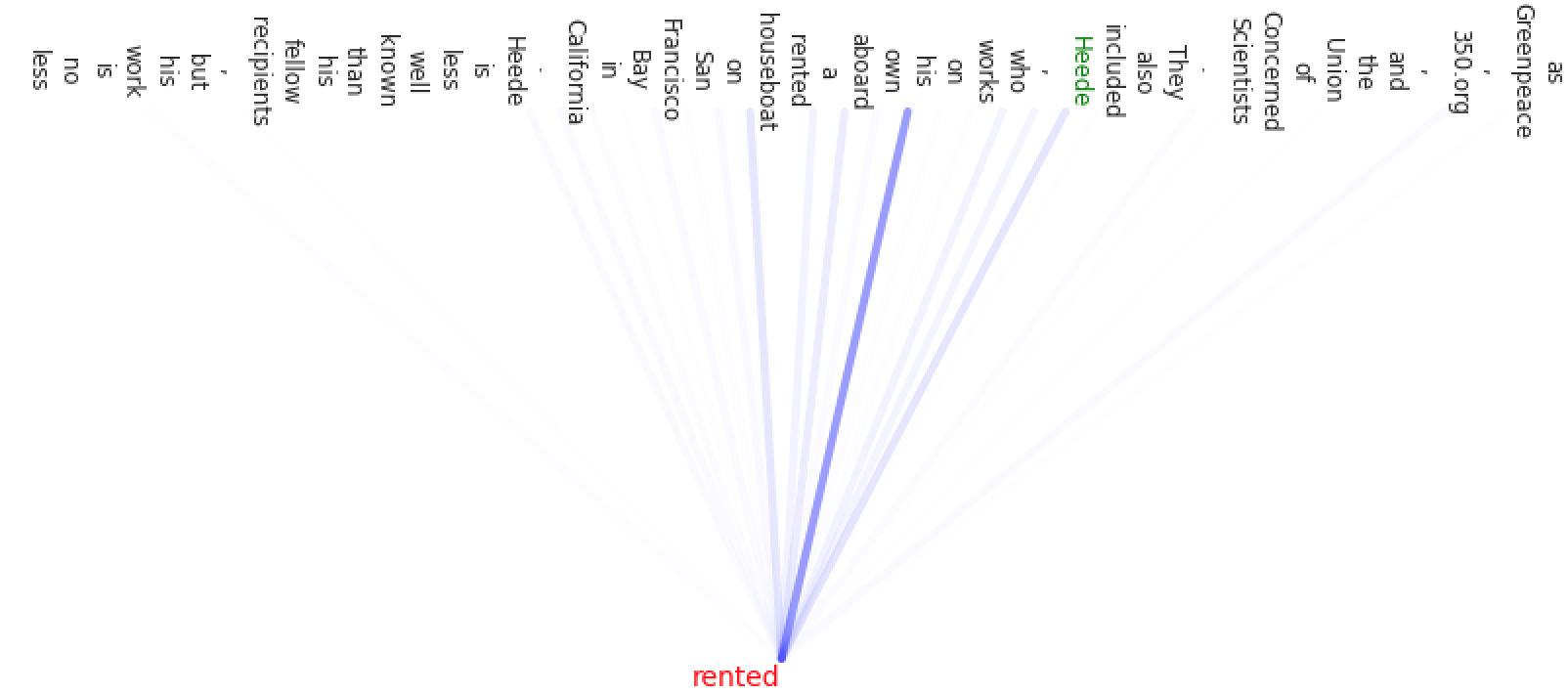}
    \caption{}
    \label{fig:beneficiary134}
\end{subfigure}
\caption{\small In (a), \textsc{BestHead} correctly picks out the \textsc{Target} of ``\textit{airstrike}" as ``\textit{Yemen}". In (b),  \textsc{BestHead} correctly picks out the \textsc{Recipient} of ``\textit{advised}" as ``\textit{companies}". In (c), the token picked is coreferent but not identical to the gold argument. Attentions are shown as \textcolor{blue}{blue} lines from trigger token, with lineweight $\propto$ value. Gold arguments are shaded {\color{green} green }.}
\end{figure} \label{fig:qualExamples}

\section{Related Work}
\label{sec:related}
A complete description of the large body of work on probing is beyond the scope of this paper. Besides those discussed earlier, other aspects studied include filler-gap dependencies \cite{wilcox2018rnn}, function word comprehension \cite{kim2019probing}, sentence-level properties \cite{adi2016fine} and negative polarity items \cite{warstadt2019investigating}. 

Probing is not limited to examining pairwise word relations or sentence properties. 
\citet{hewitt2019structural} find that BERT token representations are linearly projectable into a space where they embed constituency structure. 
Recently, \citet{sorodoc2020probing} probed transformer based language models for coreference. However, they restrict themselves to entity coreference. Furthermore, they exclude MLMs like \textsc{BERT} from their analysis. 

\citet{hewitt2019designing} raised a note of caution about classifier based probes, pointing out that probes themselves could be rich enough to learn certain phenomena even with random representations. We avoid direct classifier-based probing, thus avoiding the mentioned pitfalls.


\section{Conclusion}
\label{sec:conclusions}
We showed how BERT's attention heads have modest but well above chance ability to detect arguments for event roles. This ability is achievable either with only i) 2 parameters per role (\textsc{BestHead}) ii) 289 parameters per role (\textsc{Linear}). Furthermore, the supervision required to reach this is just $\approx 11\%$ of full training set size. Secondly, we propose a method to learning separate heads (combinations) for cross-sentence argument detection. Our experiments show that the heads so learnt have higher cross-sentence accuracy. Thirdly, we show that \textsc{Linear} performance is robust to a perturbed \textsc{NONCE} test setting with weakened lexical associations. In future, we plan to extend our probing to other event aspects like coreference and subevents.

\section*{Acknowledgments}
We thank Nikita Moghe and Hiroaki Hayashi as well as the three anonymous reviewers for their valuable feedback.

\bibliographystyle{acl_natbib}
\bibliography{emnlp2020}

\begin{thebibliography}{29}
\expandafter\ifx\csname natexlab\endcsname\relax\def\natexlab#1{#1}\fi

\bibitem[{Adi et~al.(2016)Adi, Kermany, Belinkov, Lavi, and
  Goldberg}]{adi2016fine}
Yossi Adi, Einat Kermany, Yonatan Belinkov, Ofer Lavi, and Yoav Goldberg. 2016.
\newblock {F}ine-grained analysis of sentence embeddings using auxiliary
  prediction tasks.
\newblock \emph{arXiv preprint arXiv:1608.04207}.

\bibitem[{Araki et~al.(2014)Araki, Liu, Hovy, and
  Mitamura}]{araki2014detecting}
Jun Araki, Zhengzhong Liu, Eduard~H Hovy, and Teruko Mitamura. 2014.
\newblock {D}etecting {S}ubevent {S}tructure for {E}vent {C}oreference
  {R}esolution.
\newblock In \emph{LREC}, pages 4553--4558.

\bibitem[{Bender and Friedman(2018)}]{bender2018data}
Emily~M Bender and Batya Friedman. 2018.
\newblock Data statements for natural language processing: Toward mitigating
  system bias and enabling better science.
\newblock \emph{Transactions of the Association for Computational Linguistics},
  6:587--604.

\bibitem[{Chambers and Jurafsky(2008)}]{chambers2008unsupervised}
Nathanael Chambers and Dan Jurafsky. 2008.
\newblock Unsupervised learning of narrative event chains.
\newblock In \emph{Proceedings of ACL-08: HLT}, pages 789--797.

\bibitem[{Clark et~al.(2019)Clark, Khandelwal, Levy, and
  Manning}]{clark2019does}
Kevin Clark, Urvashi Khandelwal, Omer Levy, and Christopher~D Manning. 2019.
\newblock What does {BERT} look at? an analysis of {BERT}'s attention.
\newblock \emph{arXiv preprint arXiv:1906.04341}.

\bibitem[{Devlin et~al.(2018)Devlin, Chang, Lee, and
  Toutanova}]{devlin2018bert}
Jacob Devlin, Ming-Wei Chang, Kenton Lee, and Kristina Toutanova. 2018.
\newblock Bert: Pre-training of deep bidirectional transformers for language
  understanding.
\newblock \emph{arXiv preprint arXiv:1810.04805}.

\bibitem[{Dunietz et~al.(2017)Dunietz, Levin, and
  Carbonell}]{dunietz2017because}
Jesse Dunietz, Lori Levin, and Jaime~G Carbonell. 2017.
\newblock The {BECauSE} corpus 2.0: {A}nnotating causality and overlapping
  relations.
\newblock In \emph{Proceedings of the 11th Linguistic Annotation Workshop},
  pages 95--104.

\bibitem[{Ebner et~al.(2020)Ebner, Xia, Culkin, Rawlins, and
  Van~Durme}]{ebner-etal-2020-multi}
Seth Ebner, Patrick Xia, Ryan Culkin, Kyle Rawlins, and Benjamin Van~Durme.
  2020.
\newblock \href {https://doi.org/10.18653/v1/2020.acl-main.718} {Multi-sentence
  argument linking}.
\newblock In \emph{Proceedings of the 58th Annual Meeting of the Association
  for Computational Linguistics}, pages 8057--8077, Online. Association for
  Computational Linguistics.

\bibitem[{Gulordava et~al.(2018)Gulordava, Bojanowski, Grave, Linzen, and
  Baroni}]{gulordava2018colorless}
Kristina Gulordava, Piotr Bojanowski, Edouard Grave, Tal Linzen, and Marco
  Baroni. 2018.
\newblock {C}olorless green recurrent networks dream hierarchically.
\newblock \emph{arXiv preprint arXiv:1803.11138}.

\bibitem[{Hewitt and Liang(2019)}]{hewitt2019designing}
John Hewitt and Percy Liang. 2019.
\newblock {D}esigning and interpreting probes with control tasks.
\newblock \emph{arXiv preprint arXiv:1909.03368}.

\bibitem[{Hewitt and Manning(2019)}]{hewitt2019structural}
John Hewitt and Christopher~D Manning. 2019.
\newblock A structural probe for finding syntax in word representations.
\newblock In \emph{Proceedings of the 2019 Conference of the North American
  Chapter of the Association for Computational Linguistics: Human Language
  Technologies, Volume 1 (Long and Short Papers)}, pages 4129--4138.

\bibitem[{Kang et~al.(2017)Kang, Gangal, Lu, Chen, and
  Hovy}]{kang2017detecting}
Dongyeop Kang, Varun Gangal, Ang Lu, Zheng Chen, and Eduard Hovy. 2017.
\newblock Detecting and explaining causes from text for a time series event.
\newblock In \emph{Proceedings of the 2017 Conference on Empirical Methods in
  Natural Language Processing}, pages 2758--2767.

\bibitem[{Kim et~al.(2019)Kim, Patel, Poliak, Wang, Xia, McCoy, Tenney, Ross,
  Linzen, Van~Durme et~al.}]{kim2019probing}
Najoung Kim, Roma Patel, Adam Poliak, Alex Wang, Patrick Xia, R~Thomas McCoy,
  Ian Tenney, Alexis Ross, Tal Linzen, Benjamin Van~Durme, et~al. 2019.
\newblock Probing what different {NLP} tasks teach machines about function word
  comprehension.
\newblock \emph{arXiv preprint arXiv:1904.11544}.

\bibitem[{Kingma and Ba(2014)}]{kingma2014adam}
Diederik~P Kingma and Jimmy Ba. 2014.
\newblock Adam: A method for stochastic optimization.
\newblock \emph{arXiv preprint arXiv:1412.6980}.

\bibitem[{Lakretz et~al.(2019)Lakretz, Kruszewski, Desbordes, Hupkes, Dehaene,
  and Baroni}]{lakretz2019emergence}
Yair Lakretz, German Kruszewski, Theo Desbordes, Dieuwke Hupkes, Stanislas
  Dehaene, and Marco Baroni. 2019.
\newblock The emergence of number and syntax units in {LSTM} language models.
\newblock \emph{arXiv preprint arXiv:1903.07435}.

\bibitem[{Linzen et~al.(2016)Linzen, Dupoux, and
  Goldberg}]{linzen2016assessing}
Tal Linzen, Emmanuel Dupoux, and Yoav Goldberg. 2016.
\newblock Assessing the ability of {LSTM}s to learn syntax-sensitive
  dependencies.
\newblock \emph{Transactions of the Association for Computational Linguistics},
  4:521--535.

\bibitem[{Lu and Ng(2018)}]{lu2018event}
Jing Lu and Vincent Ng. 2018.
\newblock {E}vent {C}oreference {R}esolution: {A} {S}urvey of {T}wo {D}ecades
  of {R}esearch.
\newblock In \emph{IJCAI}, pages 5479--5486.

\bibitem[{Marvin and Linzen(2018)}]{marvin2018targeted}
Rebecca Marvin and Tal Linzen. 2018.
\newblock Targeted syntactic evaluation of language models.
\newblock \emph{arXiv preprint arXiv:1808.09031}.

\bibitem[{Pennington et~al.(2014)Pennington, Socher, and
  Manning}]{pennington2014glove}
Jeffrey Pennington, Richard Socher, and Christopher~D Manning. 2014.
\newblock Glove: {G}lobal vectors for word representation.
\newblock In \emph{Proceedings of the 2014 conference on empirical methods in
  natural language processing (EMNLP)}, pages 1532--1543.

\bibitem[{Peters et~al.(2018)Peters, Neumann, Iyyer, Gardner, Clark, Lee, and
  Zettlemoyer}]{peters2018deep}
Matthew~E Peters, Mark Neumann, Mohit Iyyer, Matt Gardner, Christopher Clark,
  Kenton Lee, and Luke Zettlemoyer. 2018.
\newblock Deep contextualized word representations.
\newblock \emph{arXiv preprint arXiv:1802.05365}.

\bibitem[{Petroni et~al.(2019)Petroni, Rockt{\"a}schel, Lewis, Bakhtin, Wu,
  Miller, and Riedel}]{petroni2019language}
Fabio Petroni, Tim Rockt{\"a}schel, Patrick Lewis, Anton Bakhtin, Yuxiang Wu,
  Alexander~H Miller, and Sebastian Riedel. 2019.
\newblock Language models as knowledge bases?
\newblock \emph{arXiv preprint arXiv:1909.01066}.

\bibitem[{Sennrich et~al.(2015)Sennrich, Haddow, and
  Birch}]{sennrich2015neural}
Rico Sennrich, Barry Haddow, and Alexandra Birch. 2015.
\newblock Neural machine translation of rare words with subword units.
\newblock \emph{arXiv preprint arXiv:1508.07909}.

\bibitem[{Shwartz et~al.(2020)Shwartz, Rudinger, and Tafjord}]{shwartz2020you}
Vered Shwartz, Rachel Rudinger, and Oyvind Tafjord. 2020.
\newblock " you are grounded!": {L}atent {N}ame {A}rtifacts in {P}re-trained
  {L}anguage {M}odels.
\newblock \emph{arXiv preprint arXiv:2004.03012}.

\bibitem[{Sorodoc et~al.(2020)Sorodoc, Gulordava, and
  Boleda}]{sorodoc2020probing}
Ionut Sorodoc, Kristina Gulordava, and Gemma Boleda. 2020.
\newblock Probing for {R}eferential information in {L}anguage {M}odels.
\newblock In \emph{Proceedings of the 58th Annual Meeting of the Association
  for Computational Linguistics}, pages 4177--4189.

\bibitem[{Vashishtha et~al.(2019)Vashishtha, Van~Durme, and
  White}]{vashishtha2019fine}
Siddharth Vashishtha, Benjamin Van~Durme, and Aaron~Steven White. 2019.
\newblock Fine-grained temporal relation extraction.
\newblock \emph{arXiv preprint arXiv:1902.01390}.

\bibitem[{Vaswani et~al.(2017)Vaswani, Shazeer, Parmar, Uszkoreit, Jones,
  Gomez, Kaiser, and Polosukhin}]{vaswani2017attention}
Ashish Vaswani, Noam Shazeer, Niki Parmar, Jakob Uszkoreit, Llion Jones,
  Aidan~N Gomez, {\L}ukasz Kaiser, and Illia Polosukhin. 2017.
\newblock Attention is all you need.
\newblock In \emph{Advances in Neural Information Processing Systems}, pages
  5998--6008.

\bibitem[{Warstadt et~al.(2019)Warstadt, Cao, Grosu, Peng, Blix, Nie, Alsop,
  Bordia, Liu, Parrish et~al.}]{warstadt2019investigating}
Alex Warstadt, Yu~Cao, Ioana Grosu, Wei Peng, Hagen Blix, Yining Nie, Anna
  Alsop, Shikha Bordia, Haokun Liu, Alicia Parrish, et~al. 2019.
\newblock Investigating {B}ert's knowledge of {L}anguage: {F}ive {A}nalysis
  {M}ethods with {NPI}s.
\newblock \emph{arXiv preprint arXiv:1909.02597}.

\bibitem[{Wilcox et~al.(2018)Wilcox, Levy, Morita, and Futrell}]{wilcox2018rnn}
Ethan Wilcox, Roger Levy, Takashi Morita, and Richard Futrell. 2018.
\newblock What do {RNN} {L}anguage models learn about {F}iller-{G}ap
  {D}ependencies?
\newblock \emph{arXiv preprint arXiv:1809.00042}.

\bibitem[{Wolf et~al.(2019)Wolf, Debut, Sanh, Chaumond, Delangue, Moi, Cistac,
  Rault, Louf, Funtowicz, and Brew}]{Wolf2019HuggingFacesTS}
Thomas Wolf, Lysandre Debut, Victor Sanh, Julien Chaumond, Clement Delangue,
  Anthony Moi, Pierric Cistac, Tim Rault, R'emi Louf, Morgan Funtowicz, and
  Jamie Brew. 2019.
\newblock Huggingface's transformers: State-of-the-art natural language
  processing.
\newblock \emph{ArXiv}, abs/1910.03771.

\end{thebibliography}
\newpage
\appendix
\section{Qualitative Examples}
\label{subsec:qualExamplesAppendix}
\begin{figure*}
    \centering
    \includegraphics[width=1.8\columnwidth]{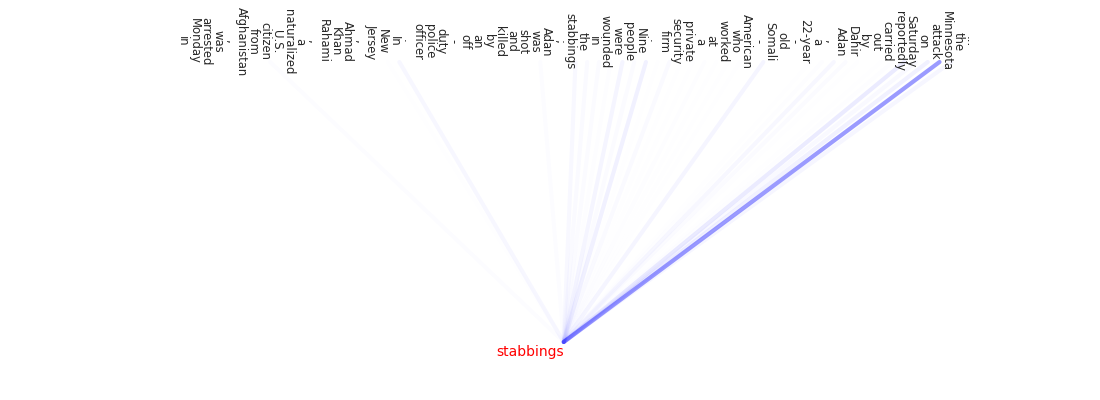}
    \caption{In this example, the head chosen by \textsc{BestHead} for the \textsc{Place} role, correctly picks out the argument as ``\textit{Minnesota}".}
    \label{fig:victim44}
    \vspace{-1ex}
\end{figure*}

\begin{figure*}
    \centering
    \includegraphics[width=1.8\columnwidth]{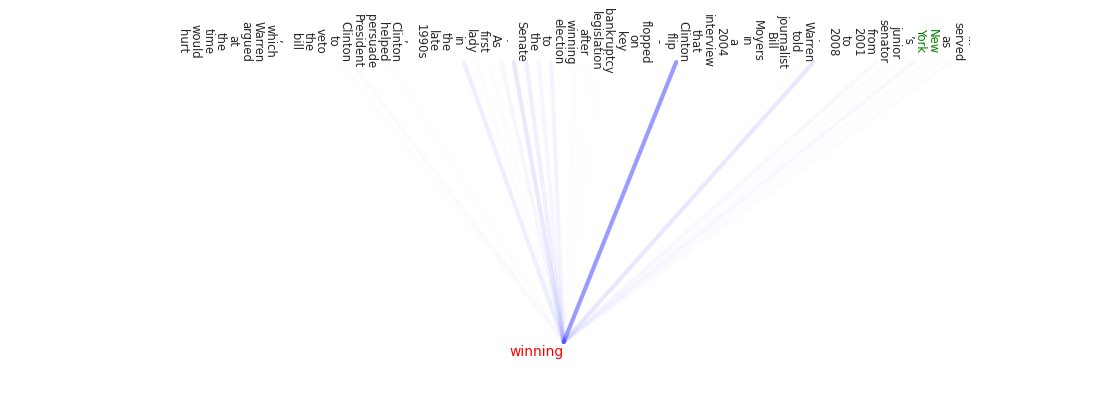}
    \caption{In this example, the head chosen by \textsc{BestHead} for the \textsc{Place} role, incorrectly picks out the argument, confusing \textit{Clinton} with \textit{New York}}
    \label{fig:place123}
    \vspace{-1ex}
\end{figure*}

\begin{figure*}
    \centering
    \includegraphics[width=1.8\columnwidth]{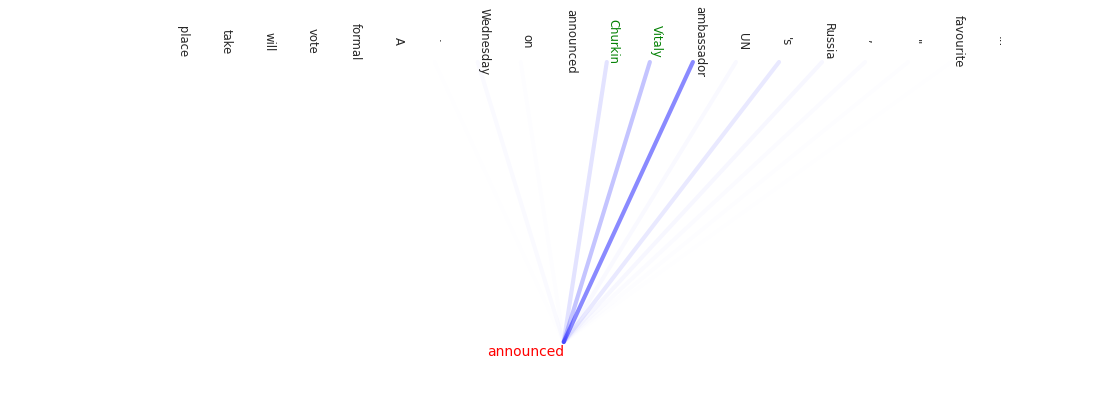}
    \caption{In this example, the head chosen by \textsc{BestHead} for the \textsc{Communicator} role, goes incorrect since it points to the adjective \textit{ambassador} of the noun phrase \textit{Vitaly Churkin}, rather than tokens in the noun phrase itself.}
    \label{fig:communicator46}
    \vspace{-1ex}
\end{figure*}

\begin{figure*}
    \centering
    \includegraphics[width=1.8\columnwidth]{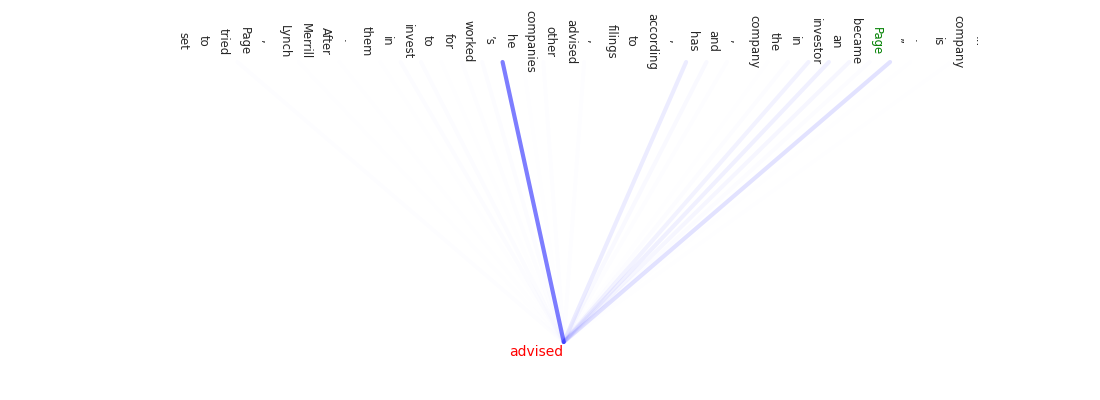}
    \caption{In this example, the head chosen by \textsc{BestHead} for the \textsc{Communicator} role, picks out the wrong coreferent (\textit{he}) rather than \textit{Page}, albeit from the correct coreference chain.}
    \label{fig:communicator51}
    \vspace{-1ex}
\end{figure*}

\begin{figure*}
    \centering
    \includegraphics[width=1.8\columnwidth]{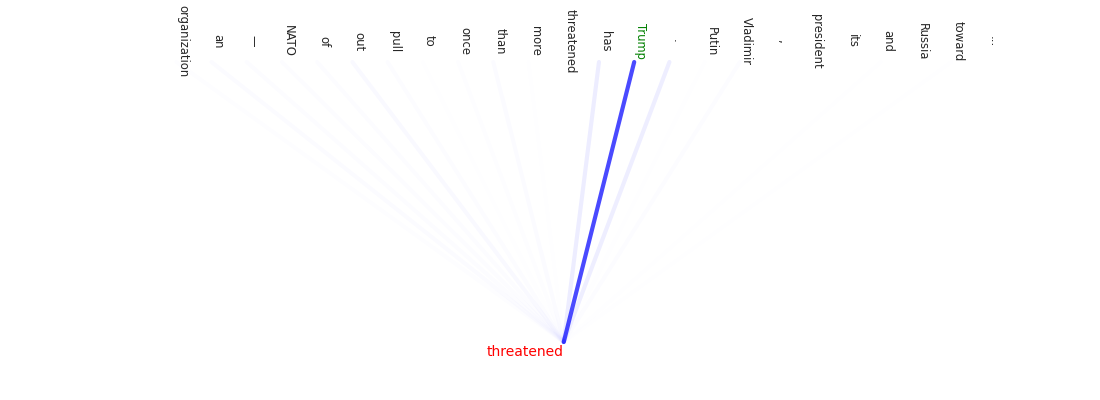}
    \caption{In this example, the head chosen by \textsc{BestHead} for the \textsc{Place} role, incorrectly picks out the argument, confusing \textit{Clinton} with \textit{New York}}
    \label{fig:communicator54}
    \vspace{-1ex}
\end{figure*}

\begin{figure*}
    \centering
    \includegraphics[width=1.8\columnwidth]{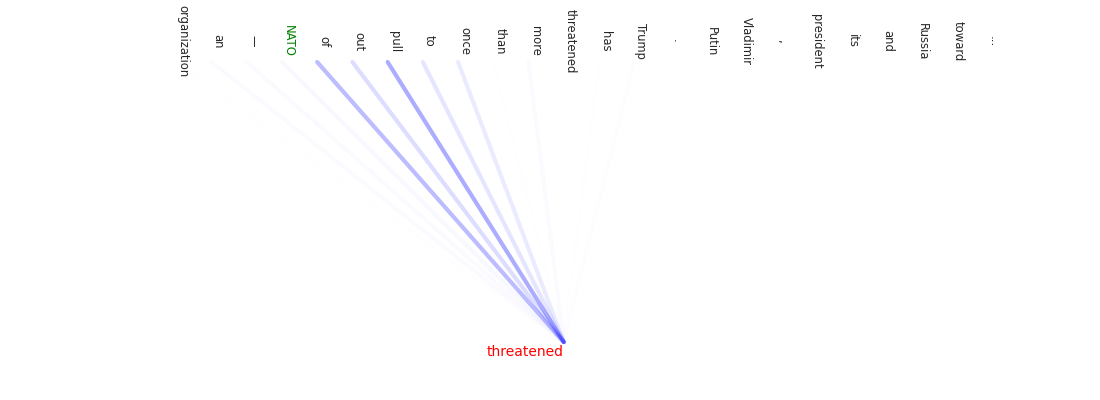}
    \caption{In this example, the head chosen by \textsc{BestHead} for the \textsc{Recipient} role, correctly picks out the gold argument \textit{NATO}}
    \label{fig:recipient54}
    \vspace{-1ex}
\end{figure*}

\begin{figure*}
    \centering
    \includegraphics[width=1.8\columnwidth]{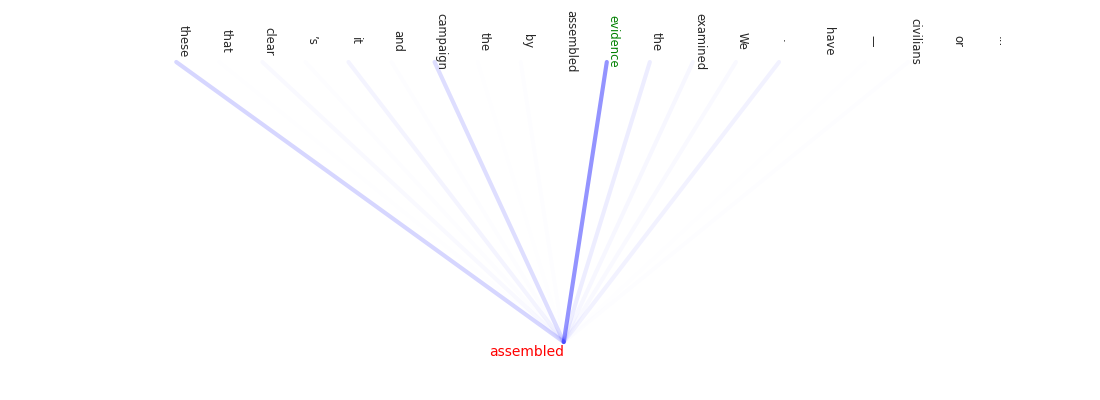}
    \caption{In this example, the head chosen by \textsc{BestHead} for the \textsc{Artifact} role, correctly picks out the gold argument \textit{evidence}}
    \label{fig:artifact96}
    \vspace{-1ex}
\end{figure*}

\begin{figure*}
    \centering
    \includegraphics[width=1.8\columnwidth]{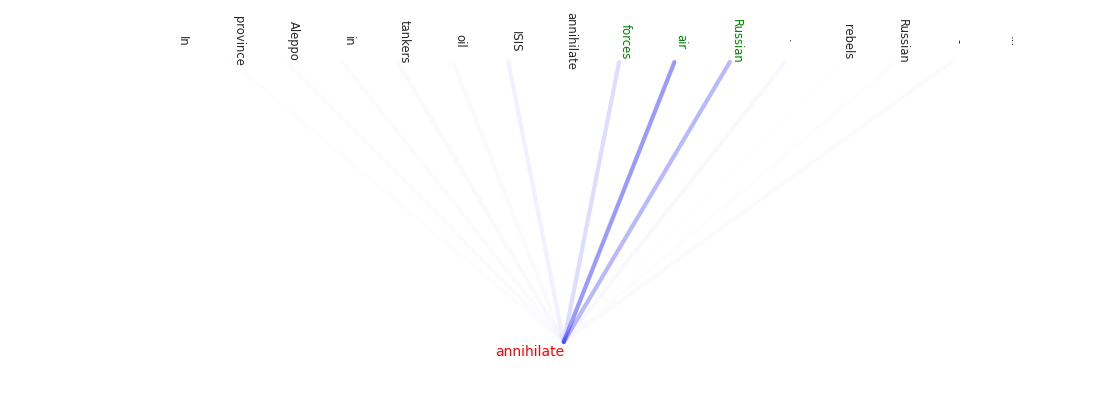}
    \caption{In this example, the head chosen by \textsc{BestHead} for the \textsc{Attacker} role, correctly picks out the argument token \textit{air} from the gold argument span \textit{Russian air force}. Furthermore, we can see that the surrounding two tokens of the gold argument span, i.e \textit{forces} and \textit{Russian} are the second and third highest attention values in this head.}
    \label{fig:attacker1101}
    \vspace{-1ex}
\end{figure*}

\begin{figure*}
    \centering
    \includegraphics[width=1.8\columnwidth]{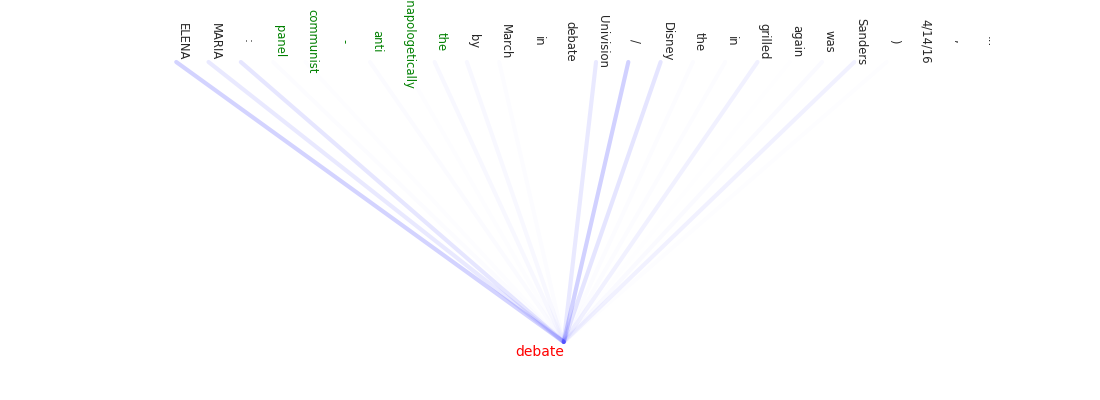}
    \caption{In this example, the head chosen by \textsc{BestHead} for the \textsc{Participant} role, incorrectly gets distracted by the reporter name \textit{Maria Elena} extraneous to the article, missing the gold argument span \textit{unapologetically anti communist panel}.}
    \label{fig:participant1}
    \vspace{-1ex}
\end{figure*}

\begin{figure*}
    \centering
    \includegraphics[width=1.8\columnwidth]{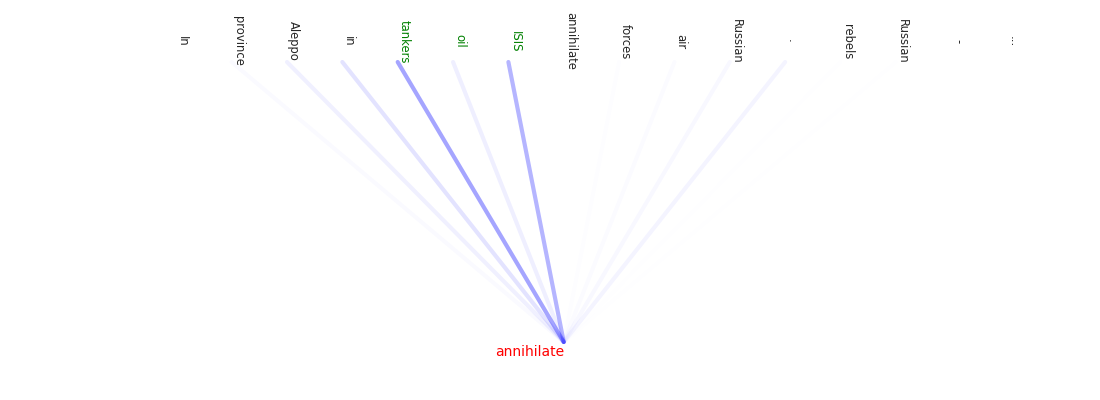}
    \caption{In this example, the head chosen by \textsc{BestHead} for the \textsc{Target} role, correctly picks out an argument token \textit{tankers} from the gold argument span \textit{ISIS oil tankers}.}
    \label{fig:target11}
    \vspace{-1ex}
\end{figure*}

\begin{figure*}
    \centering
    \includegraphics[width=1.8\columnwidth]{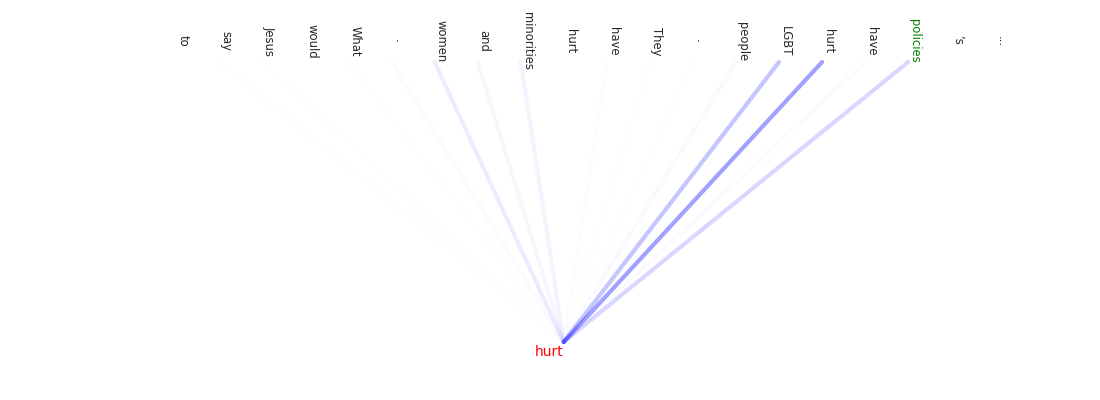}
    \caption{In this example, the head chosen by \textsc{BestHead} for the \textsc{Instrument} role, incorrectly picks out an earlier instance of the trigger \textit{hurt} instead of the gold argument token \textit{policies}.}
    \label{fig:instrument1}
    \vspace{-1ex}
\end{figure*}

Some common error types we notice for \textsc{BestHead} are:
\begin{tight_enumerate}
    \item Errors where the argument is missed due to being in another sentence. This 
    \item Errors where the argument and gold argument are in the same coreference chain, but the argument picked is a different coreferent of the gold argument and not identical. We see this in Figure 4b in Experiments. We also see this in Figure \ref{fig:communicator51}.
    \item Errors due to the same head being shared between co-occurring roles in the same example.
    \item Errors due to pointing at adjectives/adverbs of the actual noun phrase. We see this in Figure \ref{fig:communicator46} where the head points to \textit{ambassador} in \textit{ambassador Vitaly Churkin}, rather than \textit{Vitaly Churkin}, which is marked as the gold argument.
    \item Errors due to being distracted by metadata or extraneous name tokens, e.g reporter names. An example is Figure \ref{fig:participant1}.
    \item Errors due to being distracted by earlier occurrences of the same verb as the event trigger, like in Figure \ref{fig:instrument1}.
\end{tight_enumerate}
Through Figures \ref{fig:victim44} to \ref{fig:instrument1}, we present some additional qualitative examples to those in the main paper body. These include a mix of both successful argument identification as well as some failures to illustrate error types.

\begin{figure*}
    \centering
    \includegraphics[width=1.8\columnwidth]{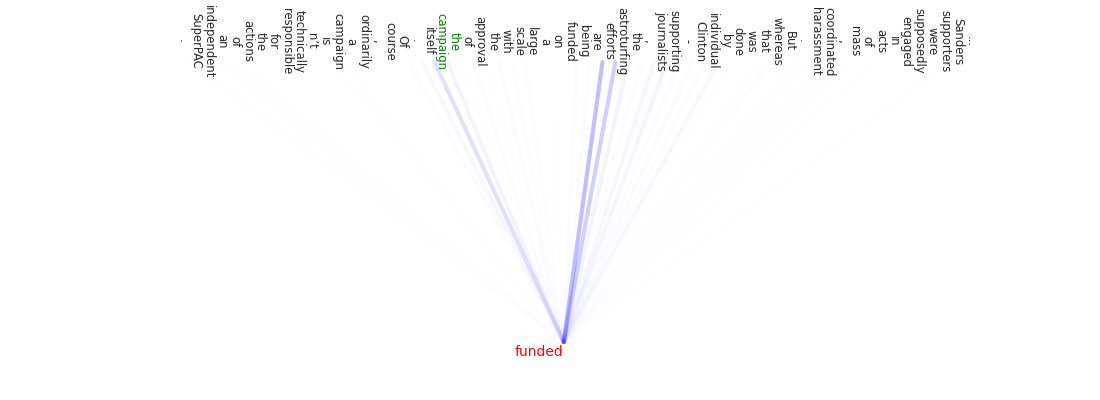}
    \caption{In this example, the head chosen by \textsc{BestHead} for the \textsc{Beneficiary} role, incorrectly picks out a subevent \textit{astroturfing efforts} referring to the full event \textit{the campaign}, which is the most explicit \textit{Beneficiary}. One may also call this a question of granularity.}
    \label{fig:beneficiary142}
    \vspace{-1ex}
\end{figure*}

\begin{figure*}
    \centering
    \includegraphics[width=1.8\columnwidth]{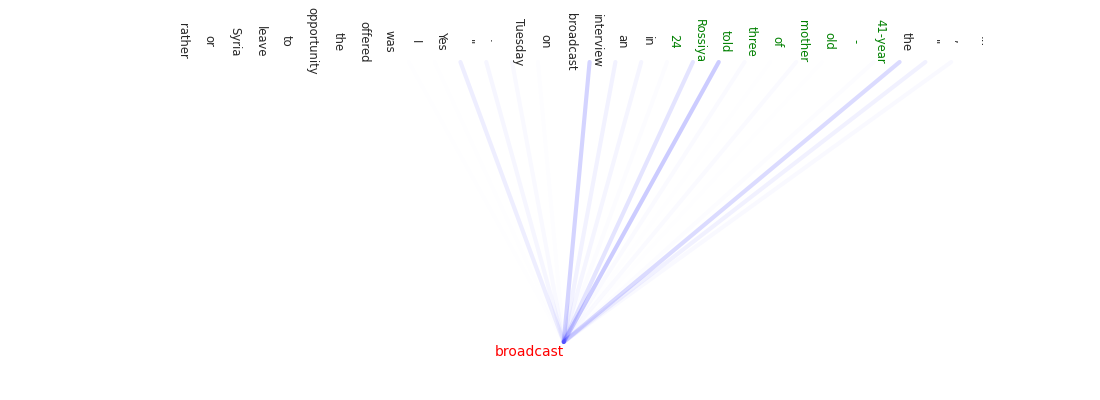}
    \caption{In this example, the head chosen by \textsc{BestHead} for the \textsc{Communicator} role, correctly picks out argument token \textit{told} from the gold argument span \textit{41-year - old mother of three told Rossiya 24}}
    \label{fig:communicator037}
    \vspace{-1ex}
\end{figure*}


\section{Other Experimental Details}

\subsection{Stop Word List}
For \S 2.5.3, the list of stop words we use is: I,you,he,she,we,they,them,our
your,mine,my,their,theirs,ours
and,or,along,with,beyond,under
forward,backward,above,below,up,down
who,what,which,how,when,
where,much,it,its,upto,until,as,
since,from,whose,whom,not

\subsection{Training Details About \textsc{Linear}}
\textsc{Linear} is trained for a maximum of $10$ epochs, using an Adam optimizer \cite{kingma2014adam} with learning rate $0.01$. Finally, the checkpoint chosen is the one with highest validation $Acc$ (as defined in \S 2.3).
\newpage
\subsection{Cross Sentence Performance: Additional Information}
In Table \ref{tab:crossSentenceOnlyApproachesComplete}, we record results for cross-sentence accuracy and the \textsc{+CSO} method for all roles
\begin{table*}
    \centering
    \small
    \begin{tabular}{l|l|l|l}
    \toprule
       Role   &  \textsc{BestHead+CSO} &  \textsc{Linear+CSO} & \textsc{Cross-Sent \%}   \\
         \midrule
         \textsc{Destination}      & 13.04 (21.43$\rightarrow$0.00) & 0 (39.28$\rightarrow$0.00) & 0 \\
        \textsc{Origin}      & 4.76 (31.82$\rightarrow$0.00) & 4.76 (56.41$\rightarrow$16.34) & 31.82 \\
        \textsc{Transporter}      & 0.00 (31.58$\rightarrow$0.00) & 0 (43.42$\rightarrow$0.00) & 15.39 \\
        \textsc{Instrument}    & 47.37 (31.37$\rightarrow$21.22) & 52.63 (25.49$\rightarrow$31.51) & 37.26  \\
        \textsc{Beneficiary}      & 0.00 (26.56$\rightarrow$0.00) & 0 (34.37$\rightarrow$0.00) & 12.50 \\
         \textsc{Attacker}      & 12.50 (33.93$\rightarrow$0.00) & 0 (46.43$\rightarrow$0.00) & 14.29 \\
        \textsc{Target}      & 0.00 (44.61$\rightarrow$21.71) & 0 (44.61$\rightarrow$0.00) & 18.47 \\
        \textsc{Giver}      & 7.14 (25.55$\rightarrow$0) & 0 (32.22$\rightarrow$0.00) & 15.56 \\
        \textsc{Victim}      & 49.99 (46.34$\rightarrow$0) & 0 (68.29$\rightarrow$0.64) & 7.32 \\
        \textsc{Artifact}      & 22.22  (50.42$\rightarrow$17.53) & 11.11 (58.82$\rightarrow$15.81) & 7.57 \\
        \textsc{Communicator}      & 19.99 (51.61$\rightarrow$0) & 9.99 (63.71$\rightarrow$15.83) & 8.07 \\
       \textsc{Participant}       & 24.99 (28.57$\rightarrow$6.37) & 29.16 (30.72$\rightarrow$6.37) & 17.14  \\
       \textsc{Recipient}      & 9.99 (40.78$\rightarrow$0) & 0.00 (44.69$\rightarrow$0) & 5.59 \\
        \textsc{Place}    & 15.31 (17.77$\rightarrow$10.18) & 30.61 (31.93$\rightarrow$9.30) & 29.52  \\        
        \bottomrule
    \end{tabular}
    \normalsize
    \caption{ Accuracies on  cross-sentence test examples using \textsc{BestHead+CSO} and \textsc{Linear+CSO}. The values $Acc_{Total}$$\rightarrow$$Acc_{Cross}$ in parentheses are the total test accuracy and cross-sentence test accuracy respectively, using the simple version of the same approach i.e \textsc{BestHead} and \textsc{Linear}.}
    \label{tab:crossSentenceOnlyApproachesComplete}
\end{table*}
\end{document}